\RequirePackage[T1]{fontenc}
\documentclass{ieeetj}
\usepackage{newunicodechar}
\newunicodechar{’}{'}
\newunicodechar{–}{--}
\newunicodechar{“}{\textquotedblleft}
\newunicodechar{”}{\textquotedblright}

\usepackage{cite}
\usepackage{amsmath,amssymb,amsfonts}
\usepackage{graphicx,color}
\usepackage{xcolor}
\usepackage{textcomp}
\usepackage{float}
\usepackage{booktabs}
\usepackage{siunitx}       
\usepackage{algorithmic}

\usepackage{times}
\usepackage{float}
\usepackage{multicol}

\usepackage{hyperref}
\hypersetup{hidelinks}
\usepackage{cleveref}

\usepackage{comment}
\crefname{figure}{Fig.}{Figs.}
\Crefname{figure}{Fig.}{Figs.}

\def\BibTeX{{\rm B\kern-.05em{\sc i\kern-.025em b}\kern-.08em
    T\kern-.1667em\lower.7ex\hbox{E}\kern-.125emX}}
\AtBeginDocument{\definecolor{tmlcncolor}{cmyk}{0.93,0.59,0.15,0.02}\definecolor{NavyBlue}{RGB}{0,86,125}}

\def\OJlogo{}
\def\seclogo{}

\begin{document}

\markboth{Terrain-Adaptive Grouser Wheel}{Griffo and Nakka}

\title{Terrain-Adaptive Grouser Wheel for Optimal Planetary Exploration: Design and Experimental Investigation
}

\author{Vincent Griffo\affilmark{1}}
\author{Yashwanth Nakka\affilmark{2}}
\affil{George W. Woodruff School of Mechanical Engineering, Georgia Institute of Technology, Atlanta, GA, USA}
\affil{Daniel Guggenheim School of Aerospace Engineering, Georgia Institute of Technology, Atlanta, GA, USA}
\authornote{AstroX, Georgia Institute of Technology.}
\hypersetup{pdftitle={Terrain-Adaptive Grouser Wheel for Optimal Planetary Exploration: Design and Experimental Investigation},pdfauthor={Vincent Griffo; Yashwanth Nakka},pdfsubject={GROVER author manuscript; submitted to IEEE Transactions on Field Robotics, revisions requested},pdfkeywords={GROVER, variable-height grousers, AstroX}}
\begin{abstract}
Planetary rovers operating in extraterrestrial environments often encounter significant mobility challenges due to varying terrain features such as gradients and granularity. While recent works in multimodal wheel design have explored adjustments in stiffness, compliance, and diameter as a means to improve terrain adaptability, full wheel grouser-adjustable designs remain largely unexplored. Grousers are a compelling feature to actuate, as granular terrains tend to require increased grouser height for improved wheel performance. As a result, we introduce GROVER, a multimodal wheel capable of continuously adjusting its grouser height for terrain adaptation. Fixed-height performance was characterized on vinyl, coarse rock, pea gravel, and sand under two packing states. Across 950 trials, the best configuration varied with terrain, reducing slip by 34.6\%--58.0\% in granular terrains and improving travel time and energy consumption by up to 77.4\%. No single grouser height minimized slip across all terrains, motivating mid-traversal reconfiguration toward lower-slip operation. Two experiments demonstrated this. First, an instructed terrain transition changed grouser height toward a known low-slip configuration during motion, with slip decreasing by $44.0\pm4.7\%$ from the transient peak across 20 trials. Second, an online adaptation routine used closed-loop slip feedback alone to search the grouser-height range for a lower-slip configuration on four terrains with previously uncharacterized granularities: three field terrains and one laboratory regolith terrain. The controller selected different heights, with settled slip 25.4\%--37.2\% lower than the pre-settled mean. Together, these results demonstrate GROVER's ability to improve mobility by varying wheel morphology during traversal toward lower-slip operating states.
\end{abstract}



\maketitle

\section{Introduction}
\IEEEPARstart{P}LANETARY rovers are the primary platforms for surface exploration and data collection on the Moon and Mars \cite{sanguino201750}. The terrains these platforms encounter are highly diverse, including loose regolith, rocky outcrops, cratered ground, and steep slopes \cite{ishigami2007terramechanics,Grotzinger2013}. While these harsh environments impose significant mobility challenges, they often contain some of the richest scientific data. For example, evidence of water ice, complex geological layering, and volatile deposits is frequently found in craters, escarpments, or lava tubes that remain inaccessible to conventional mobility systems \cite{sauro2020lava,glotch2021scientific}. These mobility challenges are further amplified in reduced gravity, which lowers the normal force available for traction and increases the risk of slippage or immobilization. A rover’s ability to overcome these conditions directly determines the quantity and quality of scientific data returned. Nevertheless, current platforms typically employ fixed-geometry wheels and conservative motion strategies to ensure reliability, often avoiding terrain features that may hold the greatest scientific interest.

\begin{figure}[!t]
    \centering
    \includegraphics[width=1\columnwidth]{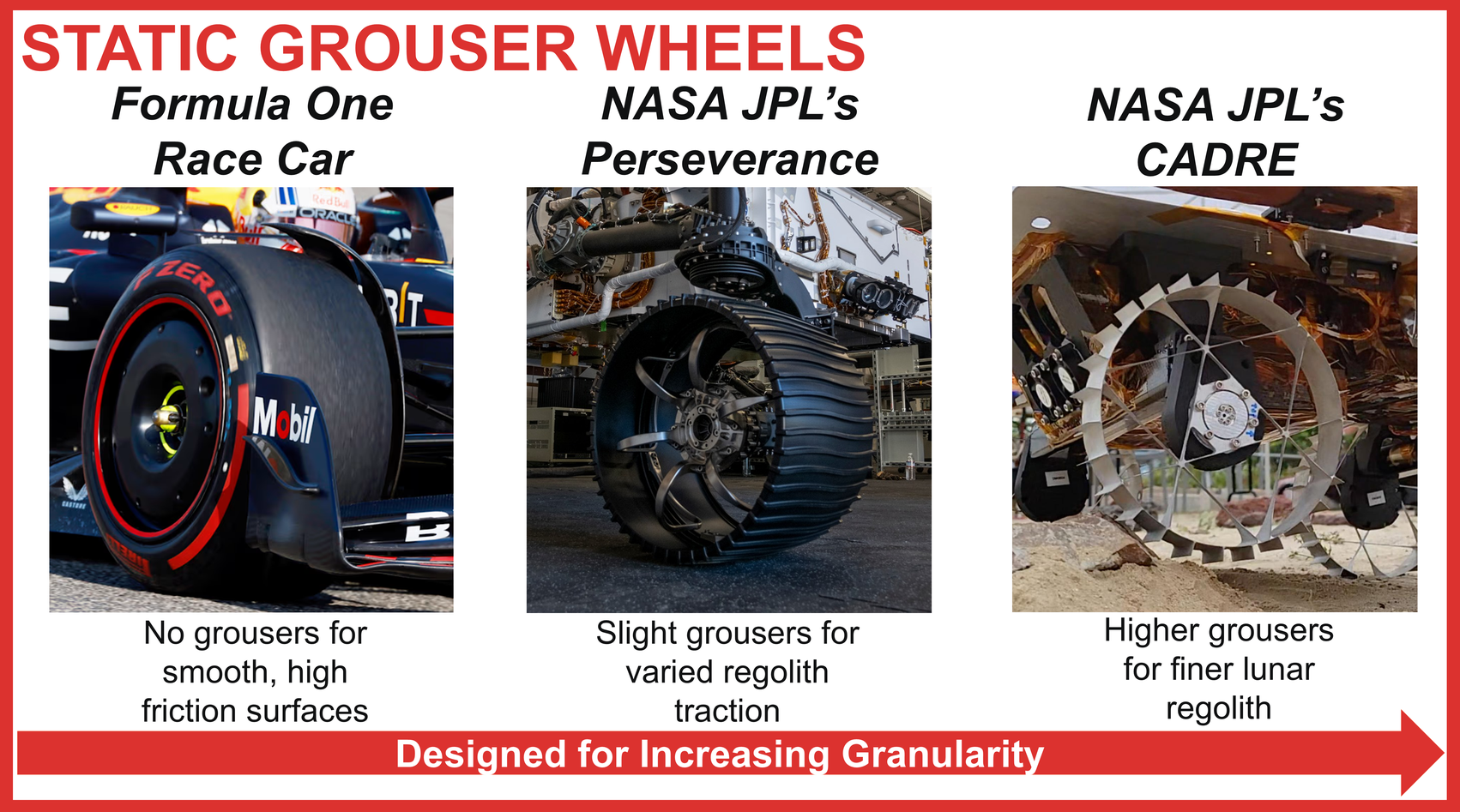}
    \caption{Static wheels, such as those on Perseverance and CADRE, use fixed grouser heights tuned to specific terrains, whereas others, like Formula~1 cars, omit grousers entirely for firm ground. Both approaches perform well only in their intended environments, limiting adaptability across varied terrains.}
    \label{fig:static}
\end{figure}

This work addresses the lack of adaptability in contemporary extraterrestrial rover designs, where wheels are typically fixed and not optimized for diverse terrain conditions. Wheel grousers, small tread elements along the wheel perimeter, highlight this limitation. Parameters such as grouser height, shape, and thickness each influence mobility differently depending on the substrate. For instance, granular media like sand or loose gravel benefit from taller grousers that penetrate the soil to increase traction and reduce slip \cite{Inotsume2019}. However, this benefit is limited, as greater grouser height also increases wheel sinkage and associated drag. Empirical observations of current platforms and their grouser heights are summarized in the Static Grouser Wheels section of \Cref{fig:static}. Grouser height must therefore be selected as a trade-off between traction and sinkage, with the optimal value depending on various soil parameters \cite{Buchele2020}. Beyond grouser design, rovers and mobility platforms more broadly benefit from multimodality, whether through varied gaits or reconfigurable wheel structures. Without adaptability, systems are forced to operate in suboptimal configurations across highly variable terrains, leading to high slip, excessive energy consumption, and reduced travel speed. In extreme cases, mobility loss can jeopardize mission objectives. A notable example is NASA's \textit{Spirit} rover, which became permanently trapped in soft Martian soil during its extended mission \cite{Callas2015}. Adaptive mobility, defined as the ability to respond to terrain variability by modifying wheel geometry, limb posture, or suspension, offers a path to extend mobility into otherwise inaccessible regions and to enable higher-value science in extreme environments. By allowing real-time adjustment to match the terrain, adaptive platforms can avoid these limitations.

\textbf{Related Work:} To improve navigation in extraterrestrial environments, recent research has increasingly investigated multimodal and bio-inspired mobility systems capable of reconfiguring their morphology to suit diverse terrains \cite{Parness2017LEMUR3, Thakker2023EELS, Aaron2011, karumanchi2017}. Prior work has also explored multimodal wheel designs in which parameters such as diameter, compliance, and stiffness are varied. \cite{Lee2024, NASA2017, Zeng2023, xu2021design, kim2013wheel, godden2024, Lee2021}. While these efforts show that transformable wheels can enhance mobility across varied surfaces, they often overlook a long-established traction-enhancing feature: grousers \cite{Inotsume2014, Nagaoka2023, Sutoh2013, sutoh2012traveling}. Studies using both simulated and constructed hardware have consistently demonstrated that wheels operating in granular media may benefit from the addition of grousers \cite{Inotsume2019,Yu2024,Takehana2025}, and that grouser geometry, particularly height and number, directly influences traction performance \cite{bauer2005experimental,ding2011experimental}. Accordingly, planetary rover wheels have almost always incorporated grousers, with recent examples including Perseverance \cite{NASA_Perseverance_Components} and CADRE \cite{NASA_CADRE}. Notably, CADRE, designed for the Moon’s finer regolith, employs taller treads than Perseverance, which operates on Mars. This contrast reflects a deliberate tailoring of grouser geometry to terrain properties. However, both rovers employ fixed-wheel designs, which limit their adaptability when encountering unexpected or suboptimal surfaces. An adjustable grouser system could extend mobility across a broader range of terrains by enabling the selection of an optimal grouser height for a given surface. While prior studies have investigated actively controlled lug or grouser architectures, as well as passively adaptive grousers, these systems are limited in their ability to provide continuous, controlled, and uniform height modulation across the entire wheel \cite{Yang2014, Nassiraei2020, Ibrahim2016}. Existing cam-based mechanisms produce non-uniform tread profiles across the wheel’s central axis, and single-lug actuation approaches regulate only one engagement element at a time rather than the entire wheel morphology. Furthermore, these systems are typically evaluated on a single terrain class and do not analyze cross-terrain adaptive performance. Platforms that claim simultaneous multi-grouser actuation lack a fully developed and experimentally validated mechanical implementation, remaining largely conceptual or simulation-based rather than demonstrated hardware systems \cite{Landers2021}. The absence of a uniform, actively controlled multi-grouser wheel architecture capable of continuous, synchronized height modulation under load remains a limitation in prior work. Existing studies also do not demonstrate how such a wheel can use this capability during traversal to respond to changing or previously uncharacterized terrain.

\begin{figure}[!t]
    \centering
    \includegraphics[width=1\columnwidth]{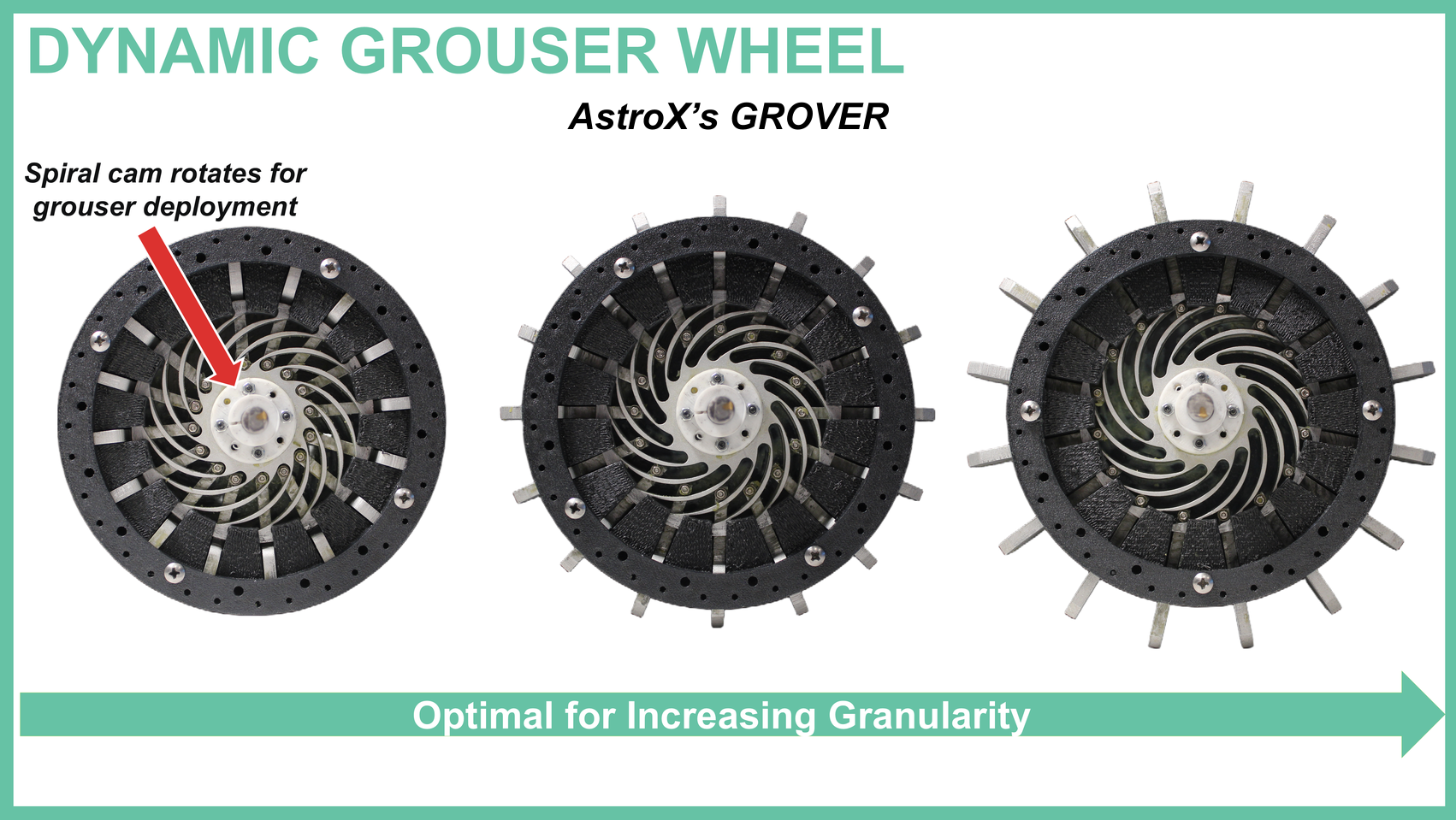}
    \caption{GROVER’s dynamic architecture employs continuous grouser adjustment (0--17.5 mm) for real-time traction optimization across diverse granular environments.}
    \label{fig:dynamic}
\end{figure}

To address this gap, the present work introduces GROVER as shown in \Cref{fig:dynamic}, a continuously variable grouser-height wheel that synchronously adjusts all 16 grousers during operation. The platform is designed around the premise that grouser height should be treated as an active mobility variable rather than a fixed wheel parameter. This is first evaluated through fixed-height testing to determine whether preferred grouser configurations vary with terrain. The resulting terrain dependence is then used to motivate two forms of mid-traversal reconfiguration: an instructed height change during a known terrain transition and an online adaptation routine that uses measured slip feedback to search for a lower-slip configuration without terrain labels or predefined height settings.

\textbf{Contributions:} This work presents the \textit{Grouser Rover Over Varied Extraterrestrial Regime (GROVER)}, a continuously variable grouser-height wheel designed to improve mobility through terrain-dependent wheel reconfiguration.

The specific contributions of this work are as follows:

\begin{enumerate}

\item \textbf{Continuously variable grouser wheel:} Design and implementation of a wheel capable of synchronized, continuous adjustment of 16 grousers through a \SI{17.5}{\milli\meter} actuation stroke. A spiral cam and internal planetary gearbox provide uniform grouser deployment during wheel operation.

\item \textbf{Closed-loop morphology control and instrumented testbed:} Development of encoder-based grouser-height estimation and closed-loop height control on a real-time, instrumented gantry testbed. Wheel and cam encoders, a \SI{5}{\micro\meter} linear encoder, and current sensing provide synchronized measurements of grouser height, slip, travel time, and drive energy during terrain interaction. This paper uses high-level sensing to estimate slip over time, as well as average slip, energy consumption, and travel time. The aim of this work is not to derive terramechanics insights, but rather to draw on concepts from the field to assess the system design.

\item \textbf{Terrain-dependent mobility characterization:} Fixed-height evaluation across 950 trials and 38 terrain-height configurations demonstrates that the best tested grouser height varies with terrain. Across granular terrains, the best configurations reduce slip by 34.6\%--58.0\% and improve travel time and energy consumption by up to 77.4\%.

\item \textbf{Mid-traversal grouser reconfiguration:} Demonstration of an instructed grouser-height change during continuous traversal across a terrain transition. Across 20 trials, reconfiguration was followed by a $44.0\pm4.7\%$ reduction from the transient slip peak.

\item \textbf{Online slip-driven adaptation:} Evaluation of autonomous grouser-height selection across four terrains with previously uncharacterized granularities: three field terrains and one laboratory regolith terrain. Using slip feedback alone, the controller selected different final heights across the four surfaces, with settled slip 25.4\%--37.2\% lower than the pre-settled mean.

\end{enumerate}

\textbf{Organization:} The remainder of this paper is organized as follows. Section II presents the preliminaries used throughout the work, including grouser-spacing criteria, wheel slip, and granular volume fraction, and relates these quantities to prior terramechanics and granular-media studies. Section III presents GROVER's mechanical design, including the spiral cam, a brief loading analysis, and the planetary gearbox. Section IV describes the experimental testbed, sensing architecture, grouser-height control, and online slip-feedback adaptation routine. Section V presents the experimental evaluation, beginning with fixed-height terrain characterization and the observed sand mobility limit, followed by instructed terrain-transition and online adaptation experiments. Section VI concludes the paper and summarizes the implications of variable grouser height for terrain-adaptive wheeled mobility.

\section{Preliminaries}
\subsection{Grouser Spacing and Slip Ratio}
Previous studies have shown that both grouser shape \cite{Nagaoka2023} and spacing \cite{Inotsume2014} significantly influence tractive performance. In~\cite{Inotsume2014}, authors empirically derived a spacing condition to reduce resistive forward soil flow, which occurs when the wheel rim contacts soil unperturbed by a preceding grouser. The resulting equation sets an upper bound on the angular spacing \((\psi)\) between grousers in terms of slip \(s\), normalized grouser height \(\hat{h}\), and normalized sinkage \(\hat{z}\), shown as

\begin{equation} 
\psi < \frac{1}{1 - s} \left( 
\sqrt{(1 + \hat{h})^2 - (1 - \hat{z})^2} - \sqrt{1 - (1 - \hat{z})^2} 
\right).
\label{eq:grouser_spacing}
\end{equation}

With expected sinkage \((z)\) and grouser height \((h)\) normalized by the wheel radius \((r)\), the angular spacing between grousers \((\psi)\) can be estimated from slip \((s)\), shown as

\begin{equation}
s = 1 - \frac{V_{\text{real}}}{V_{\text{theoretical}}}, \quad \text{where} \quad V_{\text{theoretical}} = r \omega
\label{eq:slip}
\end{equation}

and where \( r \) is the effective wheel radius in \si{\meter}, \( \omega \) is the angular velocity in \si{\radian\per\second}, and \(V_{\text{real}}\) is the measured linear velocity in \si{\meter\per\second}.

{
\subsection{Granular Volume Fraction}
Previous studies have demonstrated that the packing state of granular terrain influences locomotion performance and wheel–soil interaction behavior \cite{Goldman2013,Goldman2019}. Variations in the packing state of the same granular material can affect slip, expected sinkage, and related performance metrics. Evaluating multiple packing states therefore provides a more representative assessment of varied terrain conditions. To parameterize this effect for the sand used in this work, the solid volume fraction $\phi$ is defined as

\begin{equation}
\phi = \frac{V_s}{V},
\label{eq:vol}
\end{equation}

where $V_s$ is the volume occupied by solid particles and $V$ is the total bulk volume. Equivalently, $\phi$ may be expressed in terms of bulk density $\rho_b$ and particle density $\rho_s$ as 

\begin{equation}
\phi = \frac{\rho_b}{\rho_s}.
\label{eq:dense}
\end{equation}
}
\section{GROVER Mechanical Design}
GROVER's variable-grouser mechanism consists of two integrated mechanical subsystems: a spiral cam for synchronized grouser articulation and an internal planetary gearbox that transmits servo torque for grouser-height actuation. Wheel propulsion is provided separately by the off-axis drivetrain described below.

\subsection{Design Preliminaries}

\begin{figure}[!t]
    \centering
    \includegraphics[width=0.9\columnwidth]{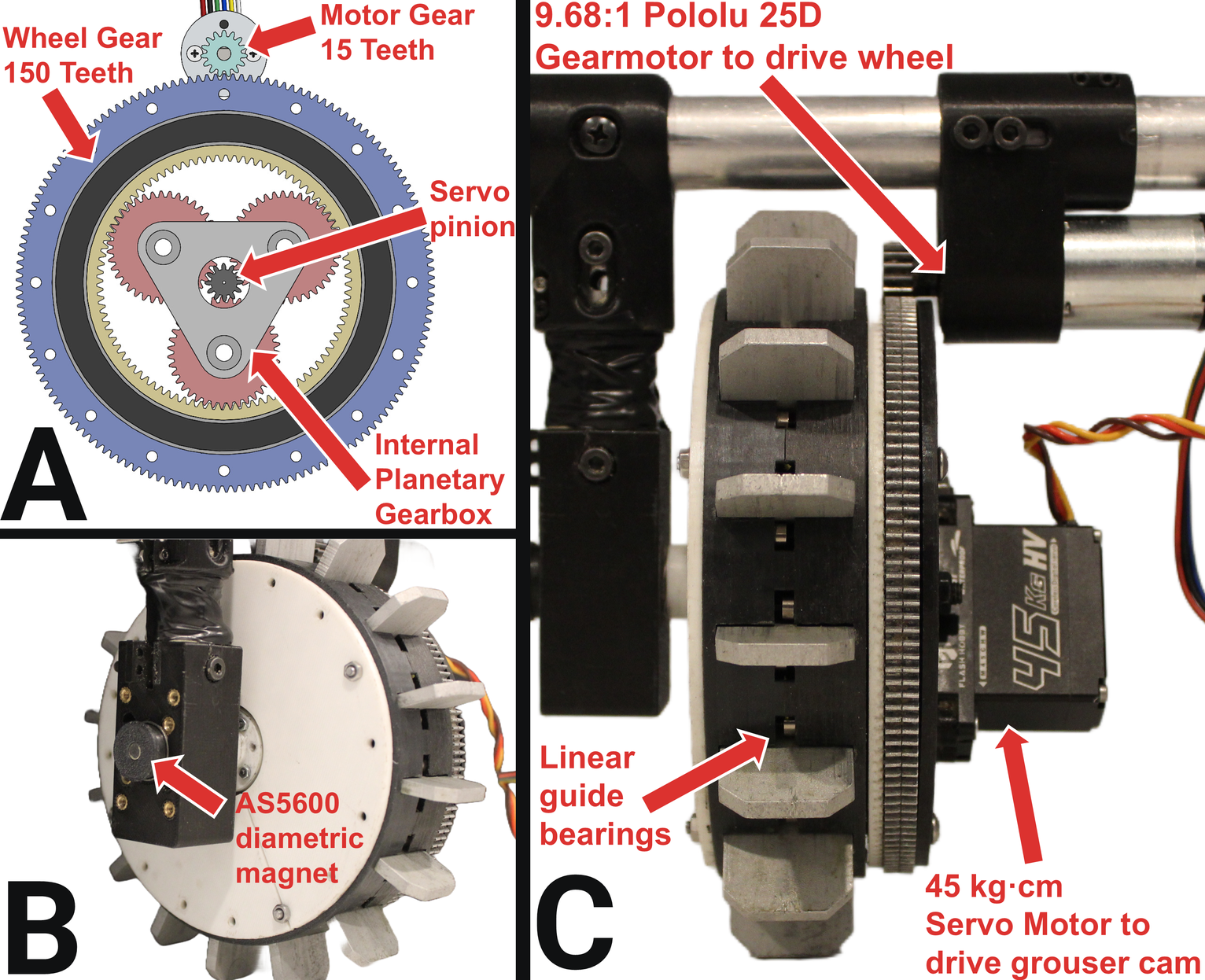}
    \caption{{(A) CAD plane view of the gear train showing the 15-tooth motor pinion driving a 150-tooth wheel gear (10:1 reduction), along with a portion of the internal planetary gearbox.
    (B) Rear view (background removed) showing the AS5600 diametric magnet used for cam angle sensing.
    (C) Side view (background removed) illustrating the \SI{4.41}{\newton\meter} servo that actuates the grouser cam, the linear guide bearings that facilitate grouser deployment, and the external 9.68:1 Pololu 25D gearmotor responsible for wheel actuation.}}
    \label{fig:Hip}
\end{figure}

GROVER is actuated by two motors: a central servo motor responsible for grouser articulation and an off-axis brushed DC motor for wheel propulsion. Because grouser deployment must rotate with the wheel frame, the servo motor is mounted along the wheel’s central axis and drives the internal planetary gear train. In contrast, the brushed DC motor is offset in a hip-mounted housing and connected via a 15-tooth pinion and a 150-tooth gear, providing a 10:1 gear ratio. The aforementioned setup has the wheel itself acting like a gear, as the 150-tooth gear and the wheel frame are kinematically synced as shown in \Cref{fig:Hip}. The designed wheel geometry was fixed at a radius of \SI{62.5}{\milli\meter} (diameter \SI{125.00}{\milli\meter}), a wheel width of \SI{42}{\milli\meter}, a main grouser width of \SI{25}{\milli\meter}, a top grouser width of \SI{15}{\milli\meter}, and a grouser chamfer angle of \SI{45}{\degree}.

\subsection{Spiral Cam}

To satisfy the spacing criteria in~\eqref{eq:grouser_spacing} while accounting for structural requirements, the spiral cam accommodates 16 uniformly spaced grousers at \SI{22.5}{\degree} intervals, producing locally thin ligaments between adjacent cam paths.

\begin{figure}[!t]
    \centering
    \includegraphics[width=1\columnwidth]{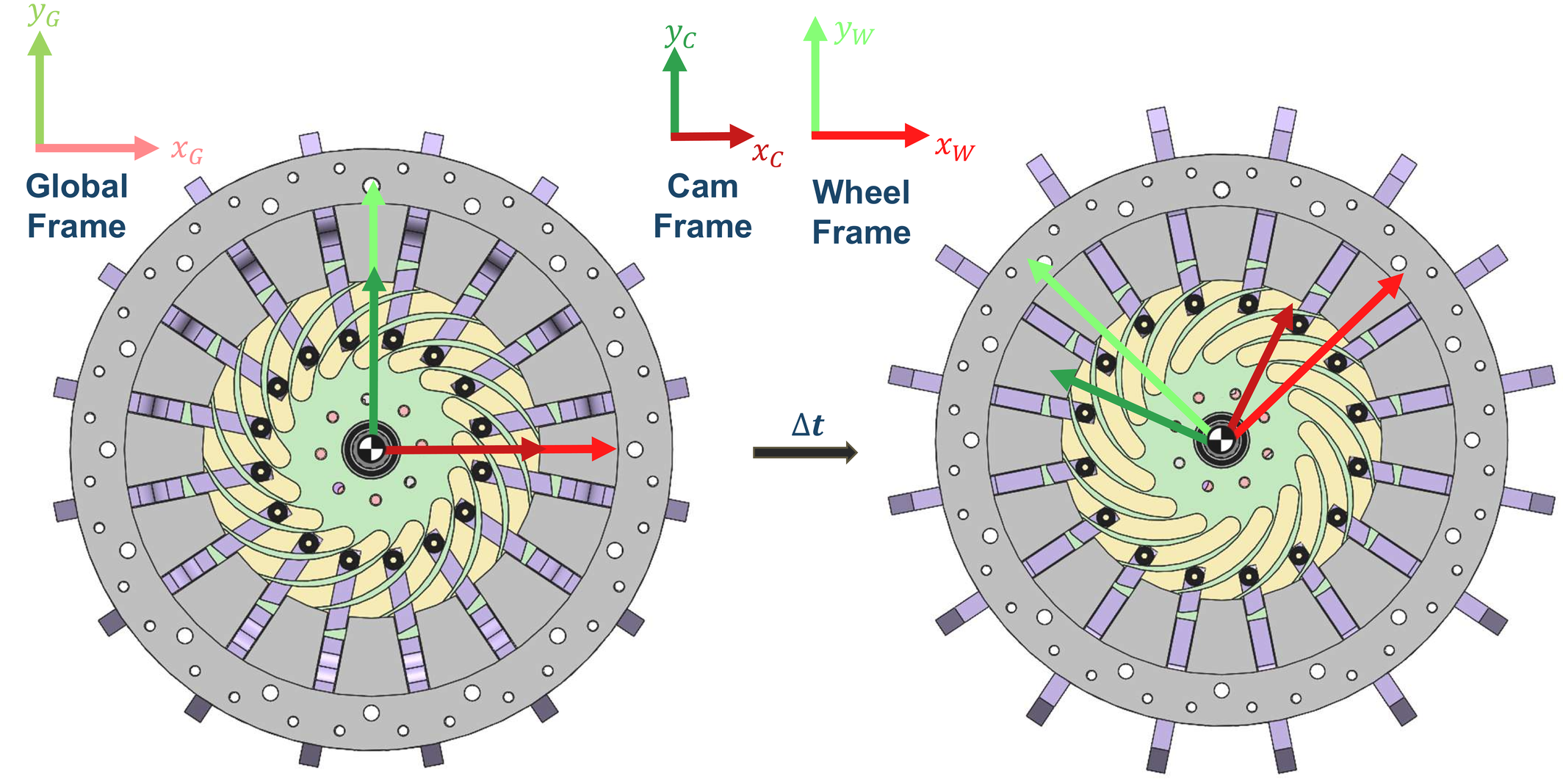}
    \caption{CAD model illustrating relative motion between the spiral cam (green) and wheel (gray) over a time interval $\Delta t$, with global, cam, and wheel reference frames.}
    \label{fig:frame}
\end{figure}

Each aluminum grouser engages the paired cam interfaces through two microbearing followers, while outer guide bearings guide radial translation. An internal bearing supports the cam independently from the wheel, allowing relative rotation and continuous grouser reconfiguration. The corresponding reference frames are shown in \Cref{fig:frame}.

\textbf{Grouser Loading:} An equivalent-static propulsion-load case assigns the complete \SI{1.23}{\kilogram} assembly weight and the drivetrain-generated tangential terrain force to one grouser:
\begin{equation}
F_r=mg\approx\SI{12.1}{\newton},
\qquad
F_t=\frac{T_{w,\max}}{r+h}.
\label{eq:grouser_loading}
\end{equation}
Using the Pololu gearmotor stall torque and external 10:1 reduction, neglecting transmission losses, gives approximately \SI{42.5}{\newton} at $h=\SI{27.5}{\milli\meter}$ and an envelope of \SI{61.2}{\newton} across the deployment range. The housing is idealized as a radial guide that permits translation while resisting transverse displacement and rotation; the cam and guide jointly balance the grouser's applied forces and moment. For the aluminum stem ($E=\SI{69}{\giga\pascal}$), with a \SI{5.5}{\milli\meter} axial width and \SI{4}{\milli\meter} depth in the wheel plane, the maximum \SI{31.41}{\milli\meter} guide-to-tip overhang produces approximately \SI{0.22}{\milli\meter} transverse deflection under the corresponding stall force. This transverse displacement is distinct from radial grouser-height error.

\textbf{Cam Loading:} The pressure angle is the acute angle between radial follower motion and the slot normal. Evaluation of the implemented slot geometry gives a maximum pressure angle conservatively rounded to $\alpha_{\max}=\SI{36.8}{\degree}$, occurring near retraction. Although this exceeds the customary \SI{30}{\degree} guideline for translating followers, the mechanism successfully deployed under the tested loads and low reconfiguration speeds using the available actuator torque. A larger wheel could accommodate more widely spaced, redesigned spiral paths, allowing the same grouser stroke with a lower average pressure angle.

The follower transmits a slot-normal reaction to the cam. Assigning the full radial load to one follower gives
\begin{equation}
F_N\leq\frac{F_r}{\cos\alpha_{\max}}
\leq\SI{15.2}{\newton}.
\label{eq:cam_load}
\end{equation}
Applying this distributed normal load throughout deployment in a linear-elastic model of the complete \SI{3}{\milli\meter}-thick aluminum cam ($E=\SI{69}{\giga\pascal}$, $\nu=0.33$), with mounting-hole boundaries fixed, predicts approximately \SI{0.068}{\milli\meter} maximum load-averaged normal displacement. Using $\alpha_{\max}$ to convert this displacement to a conservative radial-height contribution gives approximately \SI{0.085}{\milli\meter}, below \SI{0.1}{\milli\meter} under the stated model assumptions. These screening estimates exclude impact amplification and joint compliance and do not establish a complete assembly height-displacement bound. The single-grouser loading and theoretical stall torque are deliberately conservative assumptions for the applied loads and are not intended to represent typical operation.

Rolling microbearing followers limit sliding contact along the aluminum cam paths. Post-test visual inspection revealed no significant visible wear or evidence of follower seizure or binding. No measurable loss of grouser-height repeatability was observed during testing. Long-duration cyclic wear and fatigue were not evaluated in this study.

\subsection{Planetary Gearbox}

The planetary gearbox internal to GROVER provides the torque required for grouser reconfiguration. A \SI{4.41}{\newton\meter} servo drives a 12-tooth sun pinion, while the 90-tooth ring gear is coupled to the spiral cam. With the carrier constrained, the input-to-output speed ratio is
\begin{equation}
G=\frac{\omega_s}{\omega_r}=-\frac{N_r}{N_s}.
\label{eq:gear_ratio}
\end{equation}
For the implemented 12-tooth sun and 90-tooth ring gear, \eqref{eq:gear_ratio} gives $G=-7.5$, corresponding to an ideal torque multiplication of 7.5:1 before drivetrain losses.

The torque required for cam rotation depends on the applied grouser load, cam radius, local pressure angle, guide friction, and internal drivetrain losses. Pressure angle alone therefore does not define a unique stall condition. In general, the mechanism approaches stall when the total resisting cam torque approaches the torque available at the cam output. The mechanism completed its commanded grouser-deployment range under the terrain conditions evaluated in this work without observed actuator stall. An exact stall margin is not reported because gearbox efficiency, guide friction, bearing losses, and servo torque derating were not independently characterized.

\begin{figure}[!t]
    \centering
    \includegraphics[width=1\columnwidth]{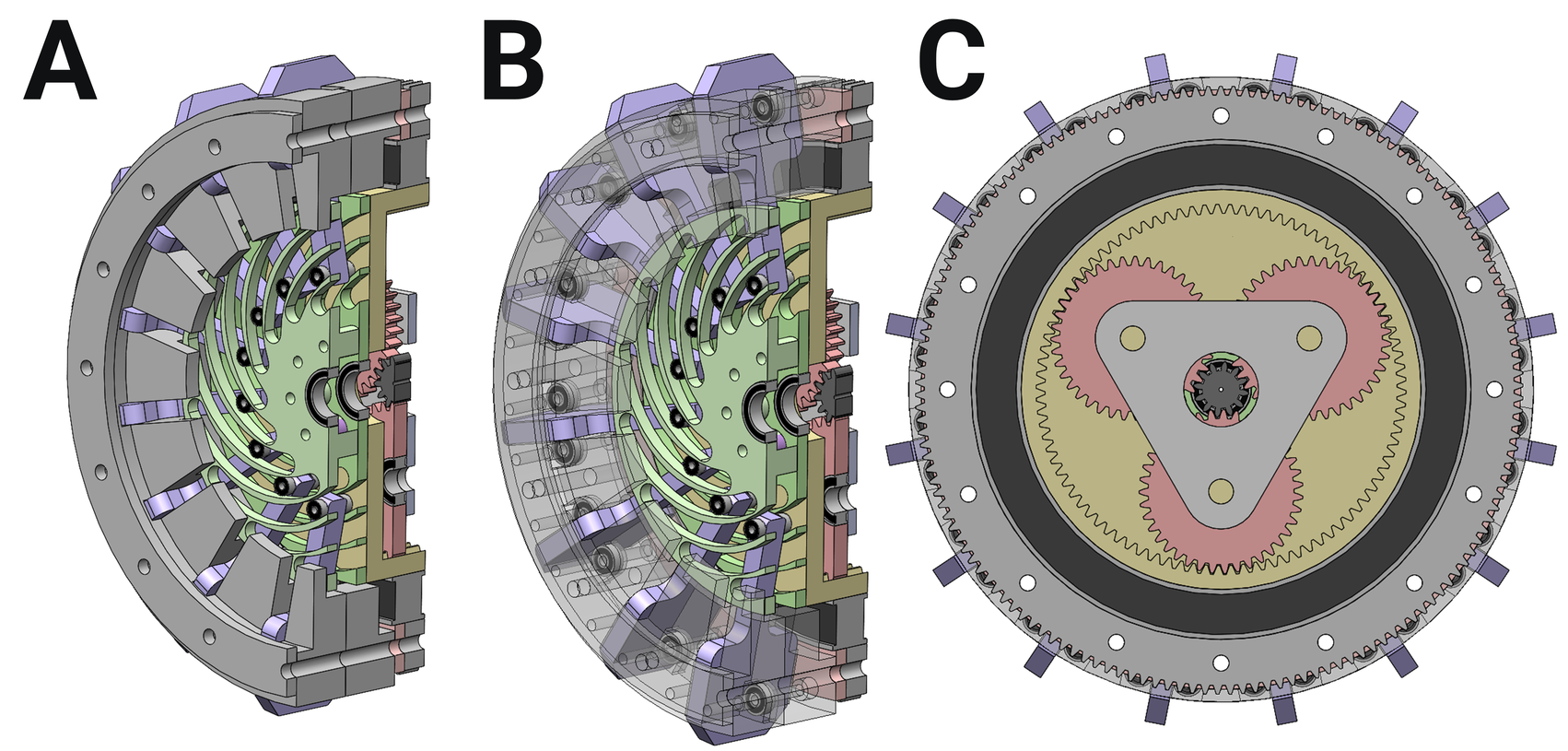}
    \caption{Cross-sectional CAD views of GROVER's internal architecture. \textbf{A)} Mid-plane section. \textbf{B)} Transparent mid-plane. \textbf{C)} Internal planetary gear train.}
    \label{fig:gearbox}
\end{figure}

The drivetrain shown in \Cref{fig:gearbox} consists of the \textbf{gray} servo pinion driving the \textbf{red} idler planet gears, which drive the \textbf{yellow} ring gear attached to the \textbf{green} spiral cam. The cam drives the follower bearings on the \textbf{purple} grousers, while the outer guide bearings guide radial translation and limit transverse grouser motion.

Measured grouser-height repeatability remained within \SI{0.1}{\milli\meter}, indicating that drivetrain backlash did not measurably limit the experimental height resolution.

A protective cover limits direct granular exposure to the spiral cam, but the translating grouser interfaces remain potential particle-ingress paths. Dust ingress was observed visually and was accompanied by audible changes during operation. This ingress did not prevent the prototype from performing its commanded reconfiguration and traversal tasks during testing; however, the present mechanism is not fully dust sealed. A space-grade implementation would therefore require improved sealing at the translating grouser interfaces and around the cam enclosure, particularly for extended operation in fine planetary regolith.

\section{Testbed \& Controls}

\subsection{Mechanical Structure}

\begin{figure}[!t]
    \centering
    \includegraphics[width=1\columnwidth]{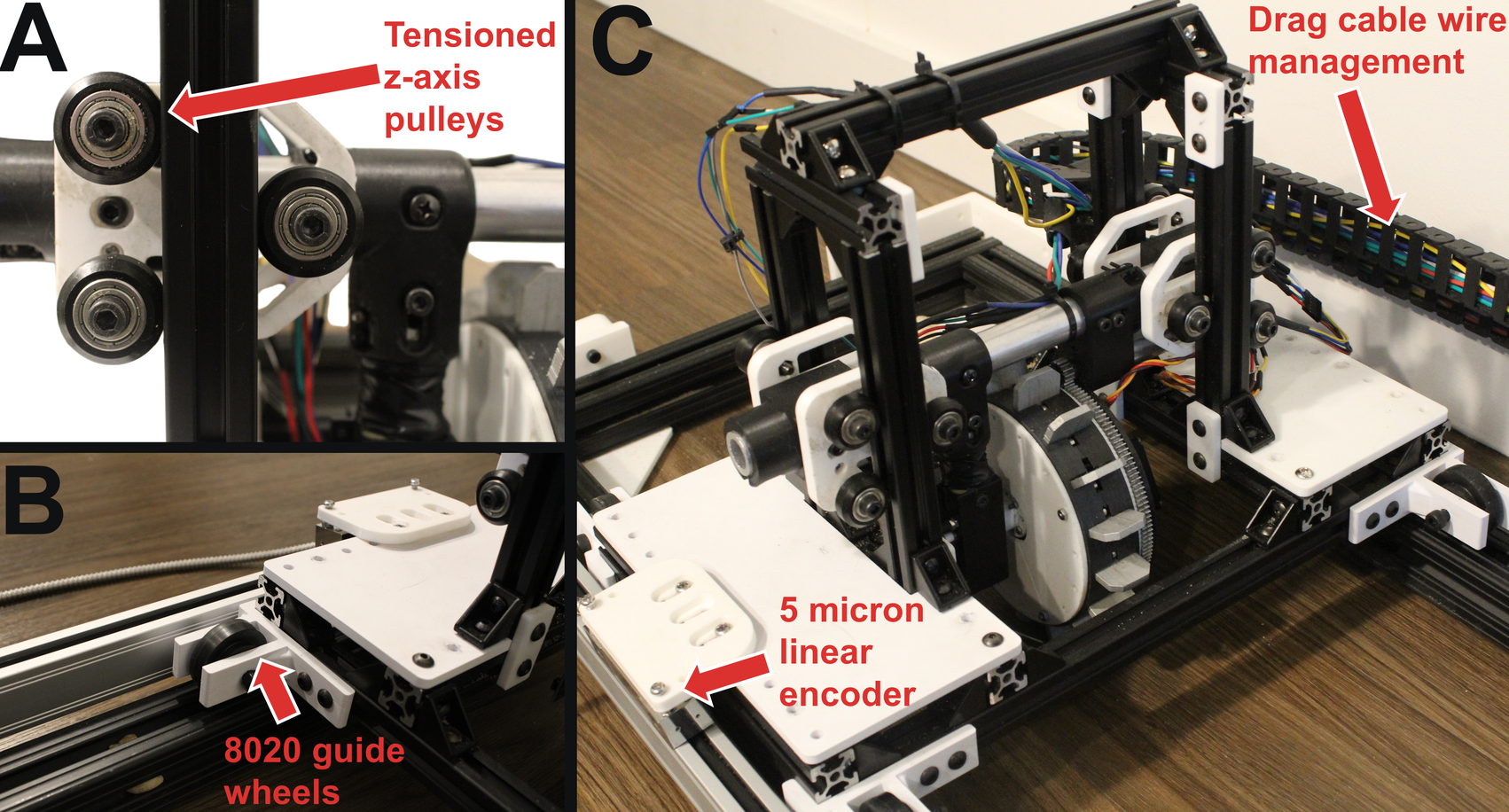}
    \caption{Annotated views of the gantry system highlighting key mechanical components. (A) Z-axis pulley assembly (background removed)  with tensioned V-wheels for vertical guidance. (B) Linear motion stage with 8020 rail and bearing interface. (C) Full gantry view showing drag chain wire routing and integrated \SI{5}{\micro\meter} linear encoder for high-precision feedback.}
    \label{fig:GantryArch}
\end{figure}

GROVER's control architecture consists of sensors located both on the wheel and on the testbed. The wheel includes two encoders and an IMU mounted at the gantry joint, which is attached to the testbed. The testbed itself is constructed from 2020 aluminum profiles for the main wheel gantry and 8020 profiles for the linear travel rail. The total stroke of the rail is \SI{432.5}{\milli\meter}, allowing sufficient distance to eliminate transient responses from the data. GROVER's gantry frame, as shown in \Cref{fig:GantryArch}, contains embedded pulleys that fit into the grooves of the 8020 extrusion. These passive pulleys ensure that GROVER is kinematically constrained to move in a single direction. GROVER, however, must be constrained not only in the direction of travel but also in height. This arrangement is necessary to emulate sinkage in more granular terrains and to allow GROVER to traverse over rocks or pebbles. To achieve this without attaching the wheel to a rigid chassis, an aluminum 2020 gantry system is used. By incorporating pulleys and eccentric nuts, a system similar to a 3D printer’s Z-axis motion is implemented and tuned for proper pulley tension. Finally, dynamic wiring associated with the translating gantry is secured using drag chains for linear motion. In the present testbed, the grouser-actuation servo is mounted to the rotating wheel and its wiring rotates with the assembly during the short-duration experimental traversals. This configuration is sufficient for the limited wheel rotations required by the testbed, but does not permit continuous wheel rotation without accumulating cable twist. For practical integration on a rover, a slip ring or equivalent rotary electrical interface would therefore be required between the stationary hip structure and the rotating wheel to transmit servo power and control signals without constraining wheel rotation.

\subsection{Electronics, Sensing, and Performance Estimation}

GROVER's sensing and electronics architecture supports grouser height measurement, slip estimation, and power/travel time monitoring. The system includes dedicated power distribution, onboard and offboard sensors, and a real-time processing architecture.

\begin{figure*}[!t]
    \centering
    \includegraphics[width=1.0\textwidth]{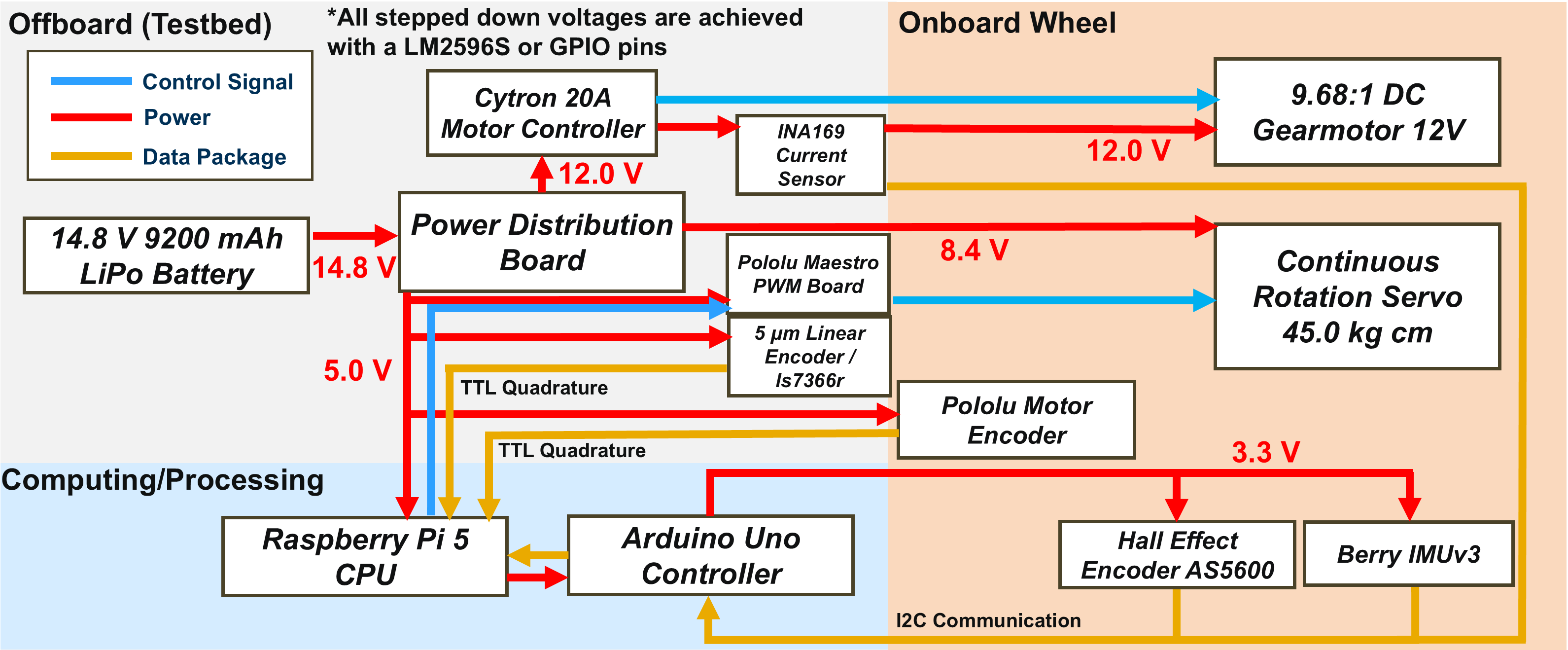}
    \caption{Testbed electronics diagram for GROVER, showing power distribution, actuation, sensing, and processing. A LiPo battery supplies regulated voltages to drive the DC gearmotor and grouser servo, with feedback from encoders, current sensing, and an IMU, while an Arduino and Raspberry Pi manage real-time control and data logging.}
    \label{fig:wiring}
\end{figure*}

\textbf{Power Distribution: } Electrically, the system is powered by a single \SI{14.8}{\volt} lithium polymer (LiPo) battery with a rated capacity of \SI{9.2}{\ampere\hour}, selected to provide sufficient energy density for sustained testing. This battery serves as the primary power source for all onboard and off-board components, including motor drivers, sensors, controllers, and embedded processors. To accommodate the system's various voltage requirements, power is distributed across three primary voltage rails: \SI{12}{\volt} for the brushed DC drive motor, \SI{8.4}{\volt} for the continuous-rotation servo actuating the spiral cam, and \SI{5}{\volt} for logic-level electronics, including the Arduino Uno and Raspberry Pi 5. Additionally, a \SI{3.3}{\volt} line from the Pi is used to interface I\textsuperscript{2}C devices, such as the Berry IMU and AS5600 magnetic encoder. The \SI{12}{\volt}, \SI{8.4}{\volt}, and \SI{5}{\volt} rails are generated using LM2596S buck converters whose outputs are manually adjusted to the required voltages. Multiple \SI{1000}{\micro\farad} capacitors stabilize the regulated rails until any voltage dip during experimentation is negligible. These converters are wired downstream from the power-distribution-board outputs. \Cref{fig:wiring} illustrates the power distribution and communication architecture across all devices.

\textbf{Sensors: } GROVER's off-board sensors include a {\SI{5}{\micro\meter} linear encoder} to measure distance traveled and an INA169 current sensor for DC motor current consumption tracking. Onboard, to enable independent estimation of wheel and cam frames, the brushed motor's encoder is paired with an AS5600 magnetic encoder. This dual-encoder setup allows for the distinction between the angular velocity of the wheel and that of the internal spiral cam, with their respective frames previously outlined in~\Cref{fig:frame}.

\textbf{Slip Estimation: } In addition to grouser actuation, encoder feedback is also central to the calculation of wheel slip. The wheel motor encoder provides a measure of angle and, once differentiated, angular velocity based on commanded motor input. A high-resolution linear encoder is simultaneously used to determine the actual translational velocity of the rover. The slip ratio is then computed using both quantities and the current wheel radius, as described by~\eqref{eq:slip}. However, due to the fine resolution of the linear encoder, standard microcontroller inputs are unable to reliably process the output. For example, at travel speeds of \SI{0.5}{\meter\per\second}, the encoder produces {\num{100000}} counts per second. These data rates exceed the interrupt handling capability of microcontrollers like the Arduino Uno or Raspberry Pi GPIO, necessitating the use of an LS7366R quadrature decoder. This dedicated interface enables lossless TTL decoding and precise position tracking.

\textbf{Energy Estimation:} To characterize wheel performance, electrical power consumption is estimated using an INA169 current sensor. For measurements performed at \SI{12}{\volt} with a constant 100\% duty cycle, instantaneous power is computed as \(P(t)=\SI{12}{\volt}\,i(t)\), where \(i(t)\) is the measured drive-motor current. Consequently, total drive energy consumption is calculated as
\begin{equation}
E=\int P(t)\,dt=\SI{12}{\volt}\int i(t)\,dt,
\label{eq:energy_cont}
\end{equation}
where the current integral is numerically evaluated using composite Simpson's rule. In discrete time, this is expressed as
\begin{equation}
E\approx\SI{12}{\volt}\frac{T_s}{3}
\sum_{j=1}^{N/2}
\Big(i[2j-2]+4i[2j-1]+i[2j]\Big),
\label{eq:energy_simpson}
\end{equation}
where $T_s$ is the sampling period, $i[k]$ is the measured current at sample $k$, and $N$ is the even number of sampling intervals, corresponding to $N+1$ samples indexed from $0$ to $N$. For sufficiently smooth current signals, Simpson's rule provides a higher-order integration approximation than rectangular or trapezoidal methods.

\begin{figure*}[!t]
    \centering
    \includegraphics[width=1\textwidth]{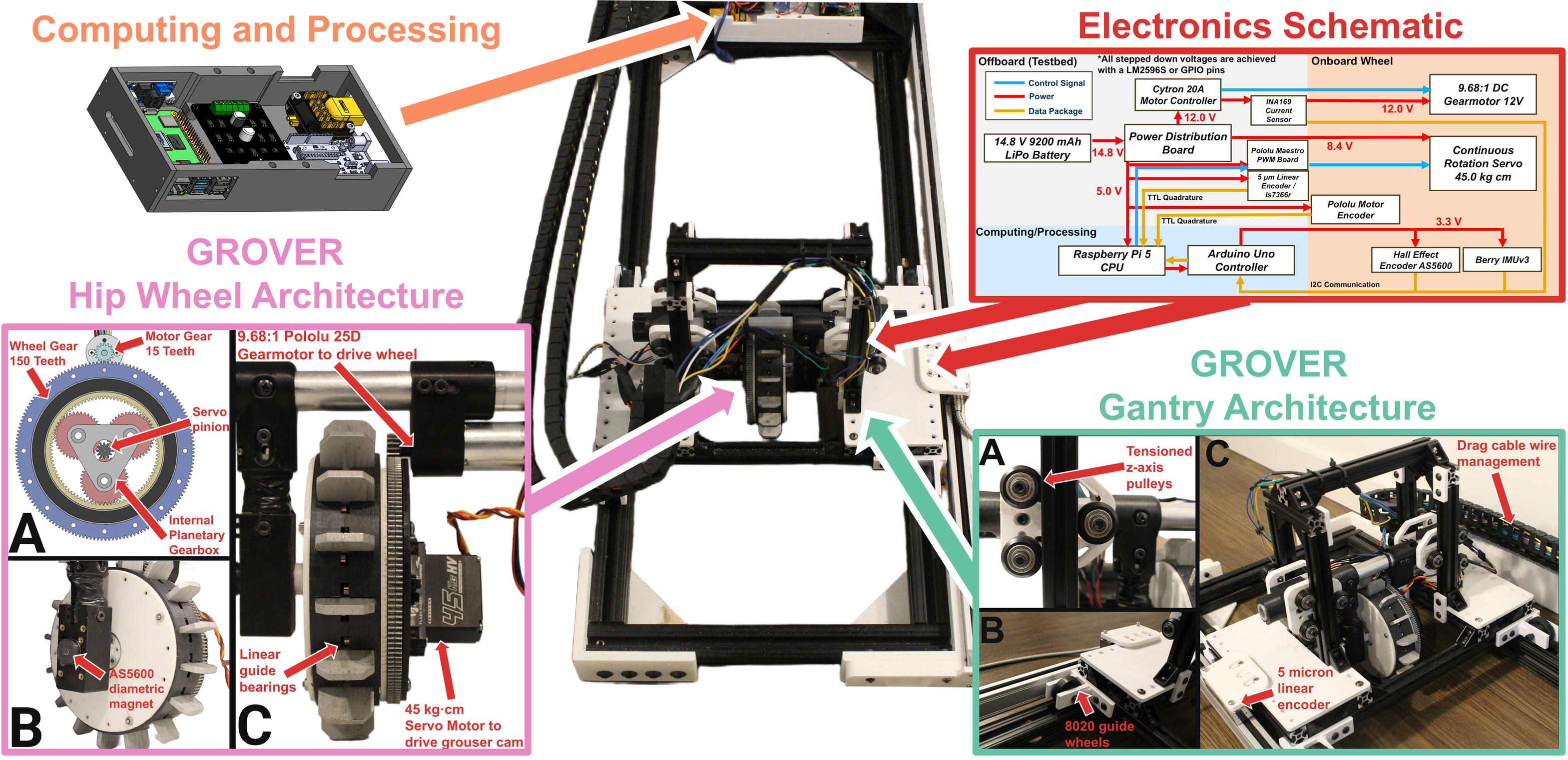}
    \caption{Overview of the GROVER testbed integrating mechanical subsystems, sensing, and control. The central image (background removed) shows the full gantry system used for terrain interaction experiments. Insets illustrate core components: (top left) onboard computing and processing unit, (top right) electronics architecture, (bottom left) GROVER's hip wheel module with internal gearing and linear encoder, and (bottom right) the gantry's Z-axis architecture for vertical actuation. Colored arrows connect each subsystem to its physical location within the integrated platform.}
    \label{fig:Testbed}
\end{figure*}

\textbf{{Real-Time Processing: }}
As shown in \Cref{fig:Testbed}, the testbed integrates all hardware and electronics used during evaluation. This setup enables GROVER to traverse diverse terrain regimes while an Arduino Uno and Raspberry Pi handle real-time sensing and processing. The I\textsuperscript{2}C AS5600 encoder is connected to the Arduino, while the Berry V3 IMU is connected to the Pi's I\textsuperscript{2}C line. Data from the current sensor, motor encoder, linear encoder, and AS5600 encoder are collected by the Arduino and communicated serially to the Pi. The Pi uses these measurements to calculate wheel slip, estimate the relative wheel--cam state, and provide the feedback signals used by the grouser-height controller described in the following subsection.

\subsection{Grouser-Height Control and Online Adaptation}
\label{sec:height_control}

The sensing architecture described previously provides both the relative cam position required to estimate grouser height and the wheel slip used to evaluate mobility performance. These measurements are used at two levels of the control architecture. First, a closed-loop controller regulates the grousers about a commanded height. Second, an online adaptation routine modifies this commanded height based on the measured slip response of the wheel.

\textbf{Grouser-Height Estimation and PID Control: }
Closed-loop grouser-height control requires a mapping between the relative rotation of the spiral cam and the resulting radial extension of the grousers. The AS5600 encoder measures the angular position of the cam, while the wheel motor encoder measures the wheel rotation. Their relative phase therefore determines the position of each follower along its corresponding cam slot and, consequently, the deployed grouser height.

The cam contains 16 geometrically identical slots distributed uniformly about the wheel at \SI{22.5}{\degree} intervals. Each slot is a rotational copy of the same three-point arc geometry and is followed by one radially constrained grouser. Because all 16 slots rotate together as a single cam relative to the wheel, the same relative wheel--cam phase produces the same radial displacement for every grouser. The following kinematic mapping is therefore derived for one reference slot and applies identically to all 16 grousers.

The physical cam slot is implemented as a three-point circular arc with center
\(C=(x_c,y_c)=(14.67,-0.69)\,\si{\milli\meter}\) and centerline radius
$R_c=\SI{24.555}{\milli\meter}$. Its Cartesian centerline is described by

\begin{equation}
(x-x_c)^2+(y-y_c)^2=R_c^2.
\label{eq:cam_arc}
\end{equation}

Because each grouser is constrained to translate radially with respect to the wheel center, the follower position can instead be expressed by the radial distance $d_f$ as a function of the relative cam phase $\theta$. Here, $\theta$ denotes the cam angle relative to the wheel, with clockwise relative rotation taken as negative:

\begin{equation}
x=-d_f\sin\theta,
\qquad
y=d_f\cos\theta,
\label{eq:polar_coordinates}
\end{equation}

Substitution of \eqref{eq:polar_coordinates} into \eqref{eq:cam_arc} gives the follower radial position directly as a function of phase. Defining

\begin{equation}
A(\theta)=x_c\sin\theta-y_c\cos\theta,
\end{equation}

the physically relevant positive radial solution is

\begin{equation}
d_f(\theta)=
-A(\theta)
+
\sqrt{
A^2(\theta)
+R_c^2-x_c^2-y_c^2
}.
\label{eq:cam_radius}
\end{equation}

The grouser extension is then obtained relative to the fully retracted cam position,

\begin{equation}
h(\theta)=d_f(\theta)-d_f(0).
\label{eq:height_mapping}
\end{equation}

\begin{figure}[!t]
    \centering
    \includegraphics[width=1\linewidth]{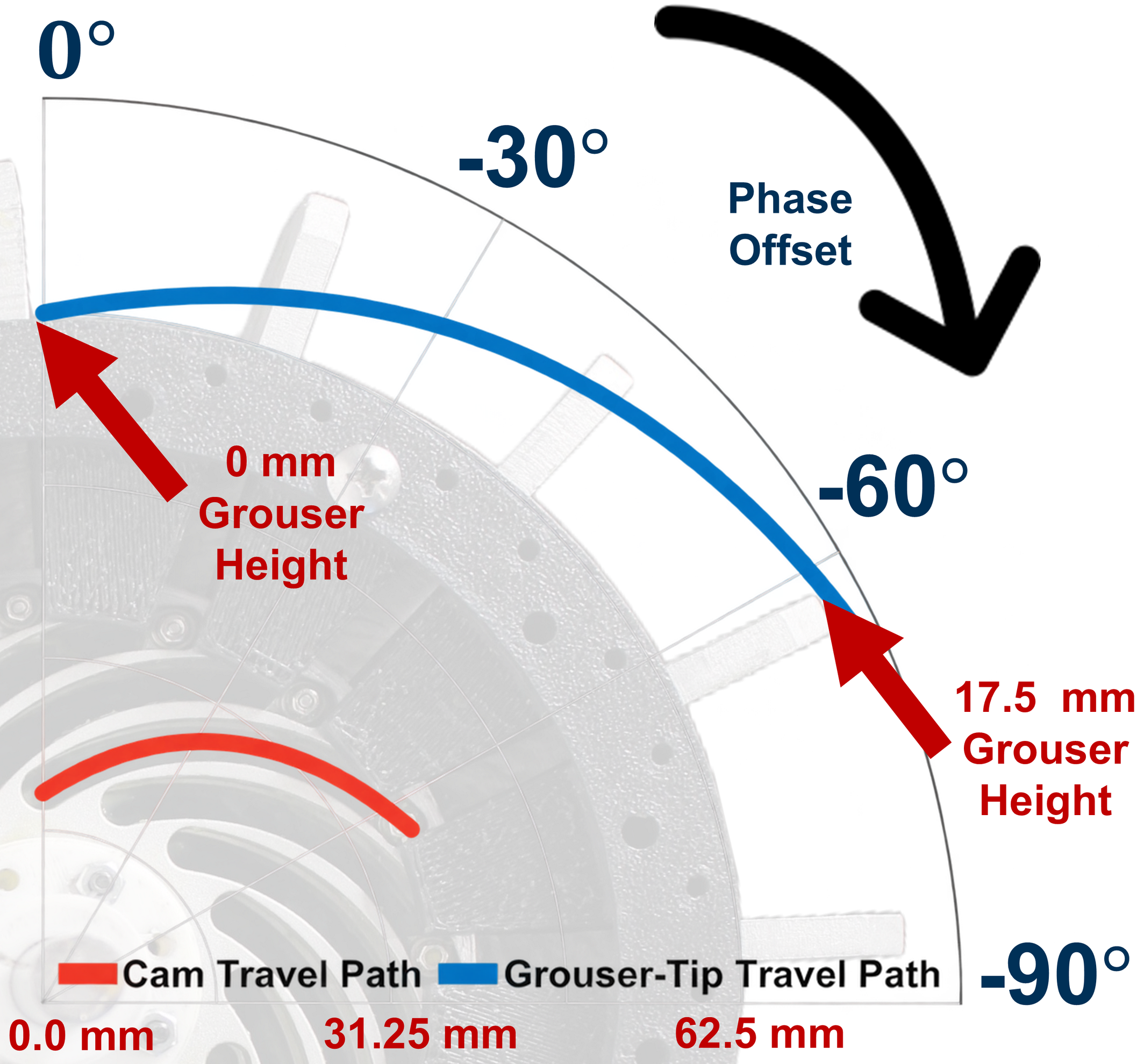}
    \caption{Polar representation of the cam and grouser-tip paths in the wheel reference frame. Relative cam phase determines the common extension of the 16 grousers, ranging from \SI{0}{\milli\meter} at \SI{0}{\degree} to approximately \SI{17.5}{\milli\meter} at \SI{-64.5}{\degree}.}
    \label{fig:spiralpath}
\end{figure}

For the implemented geometry, the follower radial distance increases from approximately
\SI{19.00}{\milli\meter} at $\theta=\SI{0}{\degree}$ to
\SI{36.50}{\milli\meter} at $\theta\approx\SI{-64.5}{\degree}$. The resulting radial displacement is therefore approximately
\SI{17.5}{\milli\meter}, corresponding to the complete grouser stroke. Since the 16 cam slots are rotationally symmetric, this relationship describes the common deployed height of all grousers for a given relative cam phase. \Cref{fig:spiralpath} shows this relationship in the wheel reference frame.

Equation~\eqref{eq:height_mapping} is evaluated using the measured relative encoder phase to obtain the grouser height $h(t)$ in real time. This closed-form geometric mapping provides the height feedback used by the grouser controller without requiring an additional interpolation step.

To regulate grouser height during operation, a PID (proportional-integral-derivative) controller is implemented. The control input is computed from the height error as

\begin{equation}
e(t)=h_d(t)-h(t),
\end{equation}

where $h_d(t)$ is the commanded grouser height and $h(t)$ is the measured height obtained from \eqref{eq:height_mapping}. The continuous-time control law is

\begin{equation}
u(t) = K_p e(t) + K_i \int_0^t e(\tau)\,d\tau + K_d \frac{de(t)}{dt},
\label{eq:PID_time}
\end{equation}

with $K_p$, $K_i$, and $K_d$ denoting proportional, integral, and derivative gains, respectively. In discrete implementation, the integral and derivative terms are approximated numerically. The integral is updated via trapezoidal summation,

\begin{equation}
I[k] = I[k-1] + \tfrac{T_s}{2}\big(e[k] + e[k-1]\big),
\label{eq:PID_integral}
\end{equation}

The derivative term is computed using a finite difference,

\begin{equation}
D[k] = \frac{e[k] - e[k-1]}{T_s + \delta_D},
\label{eq:PID_derivative}
\end{equation}

where $T_s$ is the sampling period and $\delta_D$ is a small positive regularization
constant with units of time, used to reduce the magnitude of rapid error changes. Data acquisition was sufficiently stable that higher-order filtering was unnecessary, as this would have introduced additional phase lag. Equations~\eqref{eq:PID_time}--\eqref{eq:PID_derivative} define the implemented control law, mapping height setpoints into servo actuation signals with encoder feedback. Measurements using calipers indicate that the controller achieves \SI{0.1}{\milli\meter} precision in grouser height across terrain transitions.

With angle feedback, GROVER also compensates for perturbations in grouser height. If terrain loading backdrives the spiral cam, the controller commands the servo to restore the desired relative phase. During testing, however, the cam was not observed to backdrive, as encoder measurements showed that the commanded phase offset was maintained throughout each run.

\begin{figure}[!t]
    \centering
    \includegraphics[width=1\linewidth]{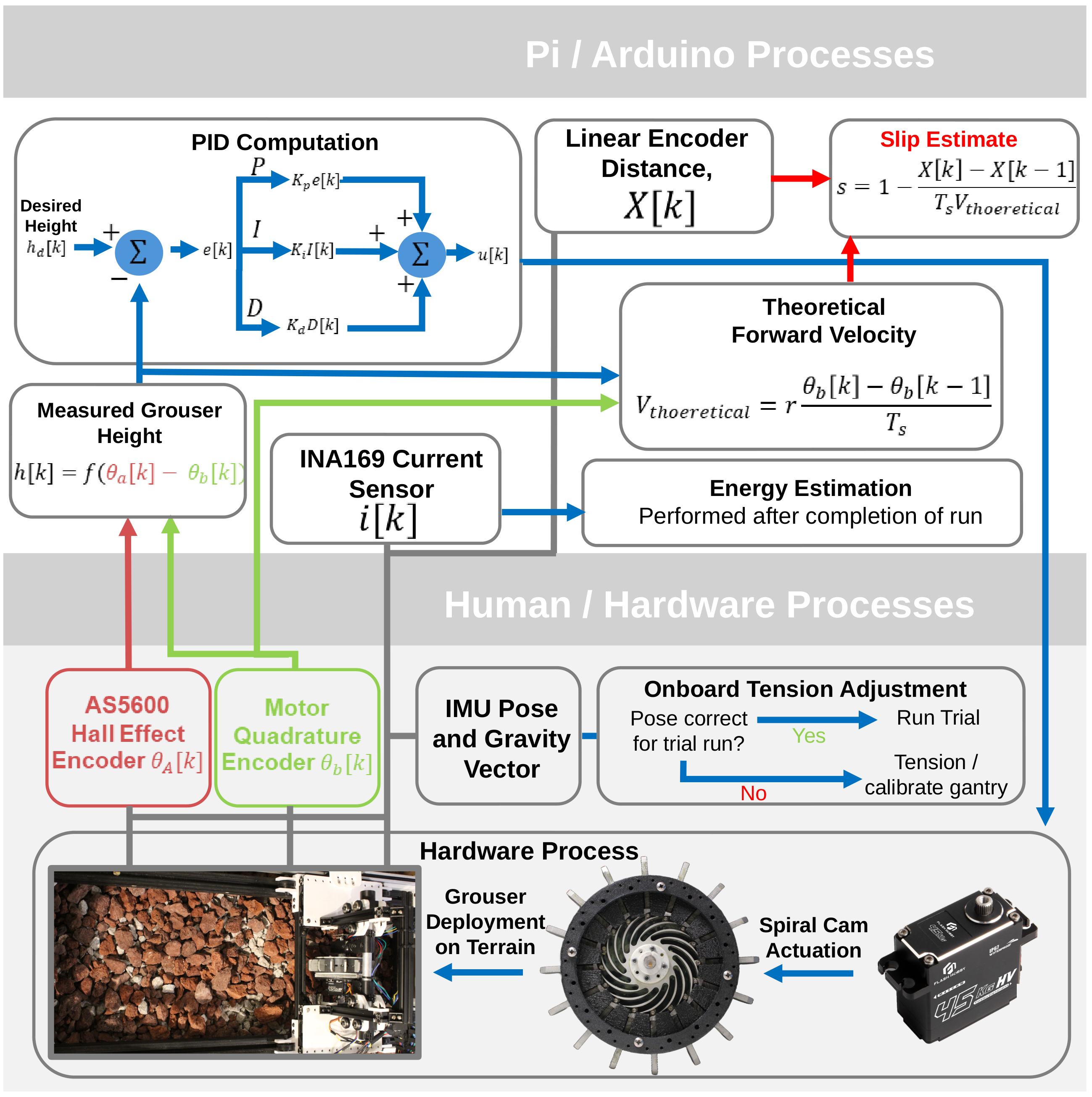}
    \caption{Closed-loop grouser height control and sensing architecture for GROVER's instrumented testbed. Wheel and cam encoders provide grouser-height feedback for PID control, while the linear encoder and motor encoder are used to estimate slip. Current sensing provides drive-energy estimates, and the IMU supports testbed pose verification and setup.}
    \label{fig:Loop}
\end{figure}

The complete closed-loop sensing and control architecture is summarized in \Cref{fig:Loop}. The measured wheel and cam angles provide grouser-height feedback through \eqref{eq:height_mapping}, while the wheel and linear encoders provide the slip estimate used for performance evaluation and subsequent online adaptation.

\textbf{Online Adaptation: }
With closed-loop grouser-height regulation established, an outer adaptation routine was developed to determine whether GROVER could adjust its grouser configuration directly from measured slip without terrain classification or a predefined height lookup. The objective is not to guarantee recovery of the globally optimal grouser height, but to demonstrate that mobility feedback can autonomously drive the mechanism toward a locally improved configuration on terrains whose granularities were not characterized in advance.

An extremum-seeking controller was initially investigated for this purpose. In outdoor testing, however, persistent and nonstationary slip fluctuations made convergence highly sensitive to filtering, excitation amplitude, and controller gains. Achieving repeatable convergence therefore required increasingly terrain-specific tuning. A simpler bidirectional search was adopted instead to provide a more robust and interpretable demonstration of slip-driven morphology adaptation.

\begin{figure*}[!t]
    \centering
    \includegraphics[width=1\linewidth]{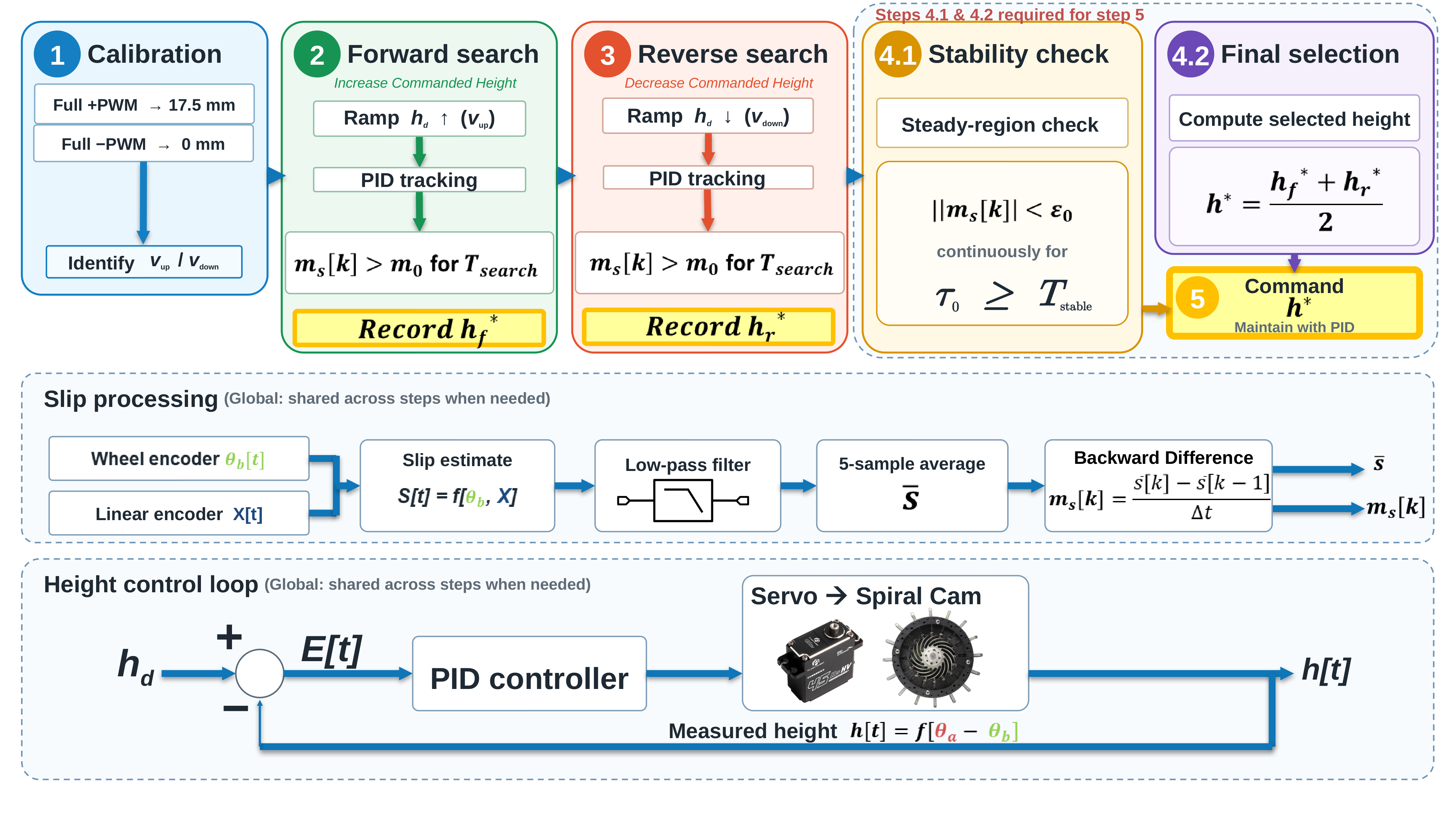}
    \caption{Online grouser-height adaptation architecture. Signals denoted by $[t]$ correspond to time-varying physical measurements and control states, whereas $[k]$ denotes the discrete index of the processed slip samples after moving-average filtering. A full-stroke calibration identifies the grouser extension and retraction rates, after which forward and reverse height sweeps are performed under PID height tracking. Slip is estimated from the wheel and linear encoder measurements, low-pass filtered, and averaged over five samples to obtain $\bar{s}[k]$. The local slip trend, $m_s[k]$, is then computed from consecutive averaged samples and used to detect sustained performance degradation during each search. Candidate minimum-slip heights from the two search directions are averaged to obtain the final selected height. During the stability check, $\tau_0$ denotes the continuously accumulated time for which $|m_s[k]|<\epsilon_0$; the selected height is accepted when $\tau_0\geq T_{\mathrm{stable}}$.}
    \label{fig:online}
\end{figure*}

The resulting architecture is shown in \Cref{fig:online} and was implemented on the Raspberry Pi. The routine consists of a full-stroke calibration, forward and reverse height searches, a stability check, and final height selection. Searching in both directions reduces sensitivity to the initial configuration, mechanism dynamics, and isolated fluctuations in the measured slip response.

At the beginning of each adaptation sequence, the grouser mechanism is driven through its available stroke in both directions. The measured deployment and retraction rates, $v_{\mathrm{up}}$ and $v_{\mathrm{down}}$, are then used to generate repeatable height ramps during the subsequent searches, accounting for differences between commanded servo motion and actual grouser deployment.

During adaptation, slip is calculated from the wheel and linear encoder measurements using~\eqref{eq:slip} and smoothed using a trailing moving average,

\begin{equation}
\bar{s}[k]
=
\frac{1}{N_f}
\sum_{i=0}^{N_f-1}s[k-i],
\label{eq:slip_filter}
\end{equation}

where $N_f$ is the filter window length. The controller operates at $f_c=\SI{100}{\hertz}$, such that $\Delta t=\SI{0.01}{\second}$. The local slip trend is calculated from consecutive moving-average estimates,

\begin{equation}
m_s[k]
=
\frac{\bar{s}[k]-\bar{s}[k-1]}{\Delta t}.
\label{eq:slip_slope}
\end{equation}

For the forward search, the desired grouser height is increased at $v_{\mathrm{up}}$ while the minimum measured slip and corresponding height are retained. Once this region has been passed, a sustained positive slope,

\begin{equation}
m_s[k] > m_0,
\label{eq:search_threshold}
\end{equation}

indicates that further extension is increasing slip. The condition must persist for $T_{\mathrm{search}}$ to reject isolated fluctuations, after which the retained minimum-slip height is recorded as $h_f^*$. The process is repeated while decreasing height at $v_{\mathrm{down}}$ to obtain the reverse-search estimate $h_r^*$.

The final selected height is

\begin{equation}
h^*=\frac{h_f^*+h_r^*}{2}.
\label{eq:optimal_height}
\end{equation}

Averaging the two estimates reduces directional bias associated with filtering, mechanism dynamics, and transient wheel--terrain interaction. It does not, however, guarantee that $h^*$ corresponds to the global minimum of the underlying slip--height relationship.

Before the selected configuration is accepted, the slip response must also satisfy

\begin{equation}
\left|m_s[k]\right| < \epsilon_0
\label{eq:stability}
\end{equation}

continuously for at least $T_{\mathrm{stable}}$. This dwell requirement prevents convergence from being declared from a single near-zero derivative during a transient response.

The search and convergence thresholds were selected empirically from short preliminary outdoor traversals and then held fixed for the reported trials. Preliminary outdoor tuning required little adjustment across the subsequently tested terrains, and the resulting parameters were used for the formal experiments. These values should therefore be interpreted as implementation-specific thresholds rather than theoretically derived optimality conditions. The resulting trial-to-trial variation in selected height is evaluated with the outdoor adaptation results in \Cref{sec:online_experiments}.

The controller parameters are summarized in \Cref{tab:online_parameters}. The extension and retraction rates are determined during the calibration stage rather than prescribed beforehand.

\begin{table}[!t]
\caption{Online grouser-height adaptation parameters.}
\label{tab:online_parameters}
\centering
\small
\begin{tabular}{lll}
\toprule
\textbf{Parameter} & \textbf{Description} & \textbf{Value} \\
\midrule
$f_c$ & Controller update frequency & {\SI{100}{\hertz}} \\
$N_f$ & Slip moving-average window & $5$ samples \\
$m_0$ & Slip-slope search threshold & {\SI{0.040}{\per\second}} \\
$T_{\mathrm{search}}$ & Search-threshold persistence & {\SI{0.25}{\second}} \\
$\epsilon_0$ & Slip-slope stability threshold & {\SI{0.070}{\per\second}} \\
$T_{\mathrm{stable}}$ & Required stability duration & {\SI{0.50}{\second}} \\
\bottomrule
\end{tabular}
\end{table}

\section{Experimental Evaluation}

The experimental evaluation is structured to first establish how grouser
height affects mobility across terrain, and then determine whether that
height can be adjusted during traversal to improve performance.
\Cref{sec:experimental_setup} describes the terrain preparation and
experimental procedures. \Cref{sec:fixed_height} characterizes slip,
energy, and travel time across fixed grouser heights, terrain types, and
sand packing states. These tests establish that mobility performance is
strongly grouser-height dependent and that the best tested height varies
with terrain. \Cref{sec:sand_limit} then examines the mobility limit
observed at excessive grouser height in sand. Finally,
\Cref{sec:terrain_transition} and \Cref{sec:online_experiments} evaluate
grouser adjustment during traversal through a commanded terrain transition
and slip-feedback adaptation across terrains with previously uncharacterized granularities,
respectively.

Unless otherwise stated, quantities reported with $\pm$ are presented as the mean $\pm$ one sample standard deviation over the corresponding measurements or trials.

\subsection{Experimental Setup and Terrain Preparation}
\label{sec:experimental_setup}

\begin{figure*}[!t]
    \centering
    \includegraphics[width=1\linewidth]{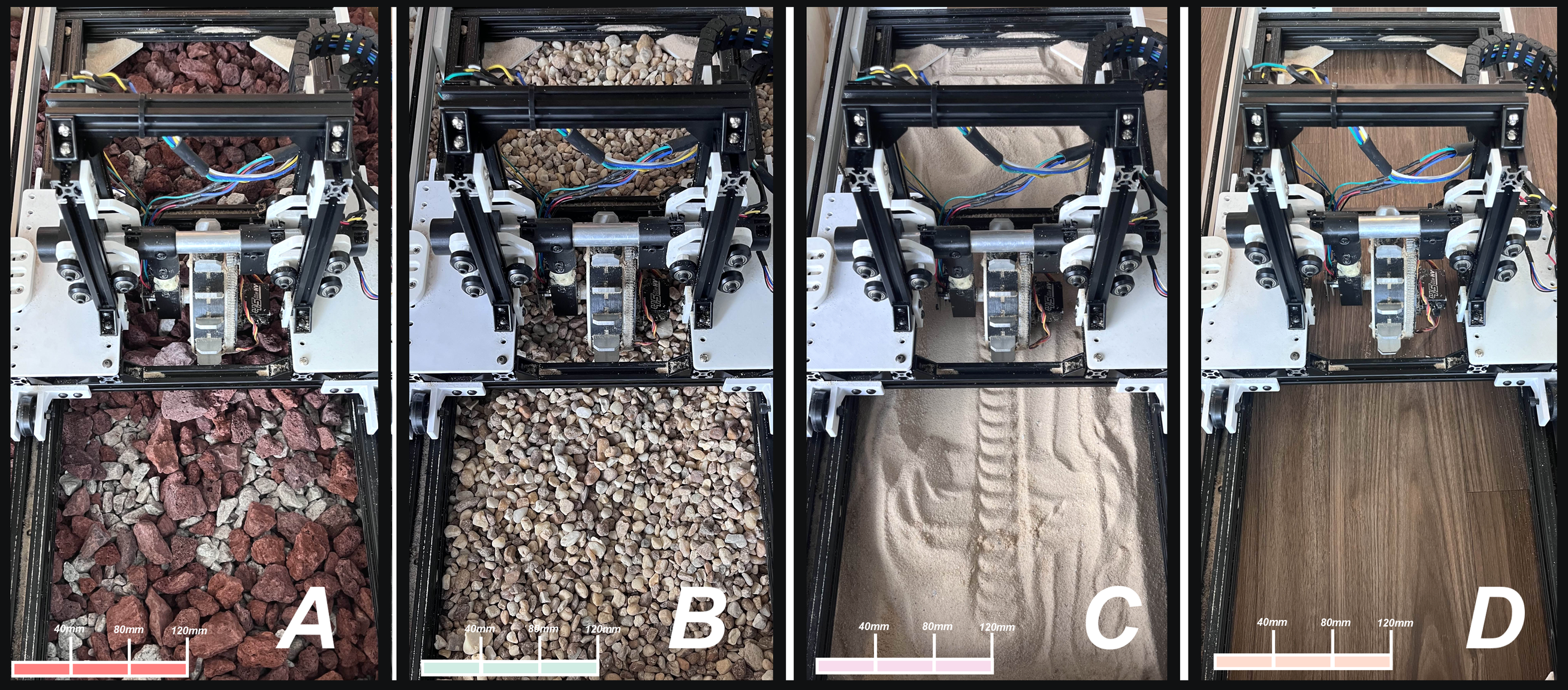}
    \caption{{Analogous field terrains used to evaluate GROVER
    across four surfaces: \textbf{A)} Coarse rock, \textbf{B)} Pea gravel,
    \textbf{C)} Sand (tested in both loose, $\phi = 0.574 \pm 0.008$, and
    densely packed, $\phi = 0.601 \pm 0.003$, states), and \textbf{D)} Vinyl
    flooring. A scale bar corresponding to the ground plane at the bottom
    of each image is shown for reference; apparent scale varies with height
    due to perspective.}}
    \label{fig:terrains}
\end{figure*}

Experiments are carried out across four distinct terrains selected to capture
a range of surface granularities. These include coarse rocks, pea gravel,
and sand, each representing progressively more unconsolidated and deformable
media. In contrast, vinyl flooring, which lacks particulate or granular
features and does not represent deformable field terrain, was used as a
non-yielding, high-slip baseline surface. This condition provides negligible
sinkage, enabling benchmarking of drivetrain and contact behavior in the
absence of granular interaction. While vinyl flooring is not intended to
replicate the friction properties of specific extraterrestrial surfaces,
such as ice that may be encountered on Saturn's moons, it serves as a
controlled reference case to understand how the adaptive grousers behave on
rigid, non-granular terrain. All terrains with their respective testbeds are
shown in \Cref{fig:terrains}.

For the sand testbed, preparation was conducted in two different ways to
obtain two packing states: loose and dense. {Quikrete Fine No.~1961
(nominal particle-size range of \SIrange{0.2}{0.6}{\milli\meter}) was first sieved using a
U.S. \#45 sieve to impose a maximum particle diameter of \SI{0.355}{\milli\meter}
for subsequent testing.} The sieved sand was then poured into the test
container at a constant rate to obtain a repeatable loose packing state.
With a particle density of {\SI{2650}{\kilogram\per\cubic\meter}}, bulk density
was determined by measuring the mass of sand placed into a container of
known volume. Repeated measurements across trials yielded a consistent
solid volume fraction of $\phi = 0.574 \pm 0.008$, which was designated
as the loose packing condition. Volume fraction was calculated using
\eqref{eq:dense}. After completing trials under loose packing, additional
experiments were conducted using a more densely packed sand state. Dense
packing was achieved by vibrating the sand containers with a motor-driven
eccentric mass to compact and settle the material. Using the same
mass--volume measurement, the dense condition produced a mean volume
fraction of $\phi = 0.601 \pm 0.003$. {These packing measurements are summarized in Table~\ref{tab:packing} and are based on 150 recorded measurements for each packing state collected during the common-range characterization at grouser heights up to and including \SI{17.5}{\milli\meter}. During the subsequent taller-grouser characterization, the same preparation procedure was used and the packing states were checked against these established volume-fraction ranges, although those checks were not retained as part of the reported packing dataset.}

Additionally, prior to testing, bulk samples from each granular terrain
were characterized using sieve analysis in accordance with standard
gradation procedures. For coarse terrains, custom 3D-printed sieve patterns
were fabricated to reproduce the desired aperture geometries. Cumulative
percent passing curves were constructed, and percentile diameters $D_p$
were obtained by interpolation. The resulting $D_p$ values are reported
in Table~\ref{tab:psd}. It is important to recognize that granular
particles are inherently three-dimensional and nonspherical, and that
their sphericity and angularity influence terramechanical response. In this
work, sieve-based percentile diameters are adopted as representative
characteristic length scales. Detailed modeling of three-dimensional
particle morphology remains beyond the scope of the present analysis.
{Consequently, the sieve-equivalent diameters are reported as simplified descriptors of the tested material gradations and do not explicitly represent higher-order morphological effects.}

\begin{table}[!t]
\centering
\renewcommand{\arraystretch}{1.1}
\caption{{{Measured loose and dense packing states for the filtered sand terrain. The reported values are based on 150 recorded measurements for each packing state collected during the common-range characterization. The same preparation protocol and established volume-fraction ranges were used during subsequent taller-grouser testing.}}}
\label{tab:packing}
\begin{tabular}{@{} lcc @{}}
\toprule
\textbf{Packing State} & \textbf{Bulk Density {(\si{\kilogram\per\cubic\meter})}} &
\textbf{Volume Fraction $\phi$} \\
\midrule
Loose  & $1520 \pm 20$ & $0.574 \pm 0.008$ \\
Dense  & $1593 \pm 8$  & $0.601 \pm 0.003$ \\
\bottomrule
\end{tabular}
\end{table}

\begin{table}[!t]
\centering
\footnotesize
\setlength{\tabcolsep}{3.0pt}
\caption{{Particle size characterization based on sieve analysis.
Percentile diameters $D_p$ were obtained by interpolation from cumulative
percent passing curves, where $D_p$ denotes the diameter at $p\%$ passing.}}
\label{tab:psd}
\renewcommand{\arraystretch}{1.05}
\begin{tabular}{lcccc}
\toprule
 & \textbf{Vinyl}
 & \textbf{Loose Sand}
 & \textbf{Pea gravel}
 & \textbf{Coarse Rock} \\

\midrule
$D_{10}$ {(\si{\milli\meter})} & -- & 0.21 & 6.8  & 22.5 \\
$D_{30}$ {(\si{\milli\meter})} & -- & 0.29 & 8.6  & 29.6 \\
$D_{50}$ {(\si{\milli\meter})} & -- & 0.33 & 9.7  & 35.1 \\
$D_{60}$ {(\si{\milli\meter})} & -- & 0.34 & 10.6 & 38.4 \\
$D_{90}$ {(\si{\milli\meter})} & -- & 0.35 & 12.5 & 46.3 \\
\bottomrule
\end{tabular}
\end{table}

Each terrain was prepared within a contained enclosure, and a consistent
surface profile was maintained across trials using a custom raking jig.
This tool uniformly reset the surface height prior to each run, eliminating
variability introduced by previous traversals and ensuring repeatability
between trials. For sand terrains, the described volume-fraction preparation
procedures were repeatedly applied and monitored to maintain consistent
packing states. Re-leveling and resetting the terrain between trials
prevented bias introduced by residual artifacts from earlier traversals.

Control variables were terrain and grouser height. To provide a common
comparison across all terrain states, six evenly spaced grouser heights were
tested: {\SIlist{0.0;3.5;7.0;10.5;14.0;17.5}{\milli\meter}}.
With four terrains, one evaluated under two packing states, this produced
30 terrain-height configurations. Wheel radius, mass, grouser count, and
grouser geometry were held constant across this common height range.
Twenty-five trials were performed for each configuration, giving 750 trials across the common height range. An additional 200 fixed-height trials were conducted at larger grouser heights to extend the characterized range and determine whether the observed terrain-dependent trends persisted or reversed beyond \SI{17.5}{\milli\meter}. The complete fixed-height characterization therefore comprised 950 trials.

All fixed-height traversals used a constant {\SI{12}{\volt}} wheel-motor input. The
wheel-and-hip assembly had a measured mass of approximately
\SI{1.23}{\kilogram}, corresponding to a static normal load of approximately
\SI{12.1}{\newton}. Performance was evaluated over the
\SI{432.5}{\milli\meter} usable traversal distance, with the first
\SI{0.3}{\second} excluded from analysis to remove the startup transient.

\subsection{Fixed-Height Terrain Characterization}
\label{sec:fixed_height}

The fixed-height experiments were used to determine whether mobility
performance depends on grouser height and whether the preferred
configuration changes with terrain. This provides the basis for active
adaptation: if different terrains favor different grouser heights, changing
wheel geometry during traversal can provide a mobility advantage over a
single fixed configuration. Slip, energy consumption, and travel time were
therefore evaluated across the tested grouser-height range for each terrain
condition.

Repeated-trial performance was consistent across the common height range,
with slip standard deviations averaging 0.0181 for pea gravel, 0.0403 for
vinyl, 0.0079 for coarse rock, 0.0227 for loose sand, and 0.0228 for
densely packed sand.

The {\SIrange{0.0}{17.5}{\milli\meter}} characterization
captures clear interior performance minima for vinyl, pea gravel, and coarse
rock. In both sand packing states, however, slip continues to decrease
through \SI{17.5}{\milli\meter}. A larger grouser-height range was therefore
used to determine whether the sand response continued to improve or reversed
at greater heights.

\begin{figure}[!t]
    \centering
    \includegraphics[width=1\columnwidth]{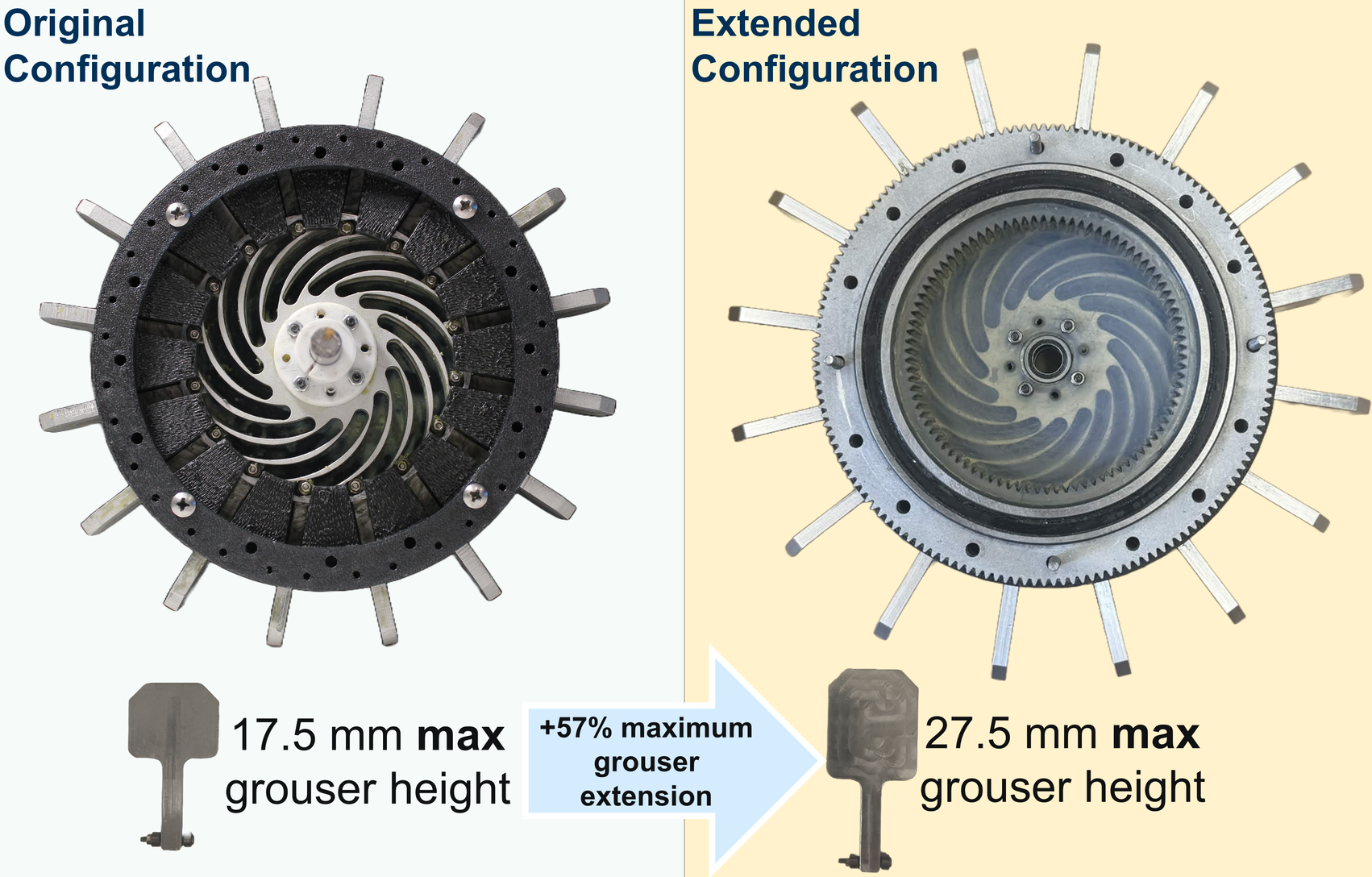}
    \caption{Original \SI{17.5}{\milli\meter} (left) and
    Extended \SI{27.5}{\milli\meter} (right) maximum-height grouser
    configurations. The latter increases the maximum exposed grouser height
    by 57\%.}
    \label{fig:newGrouser}
\end{figure}

Two interchangeable hardware configurations were used, with maximum exposed
grouser heights of \SI{17.5}{\milli\meter} and
\SI{27.5}{\milli\meter}. As shown in \Cref{fig:newGrouser}, the
\SI{27.5}{\milli\meter} configuration increases the maximum exposed height
by \SI{10}{\milli\meter}, or 57\%. Larger motor and wheel-hip clamps were
fabricated to accommodate the longer grousers. For grouser heights exceeding
approximately \SI{25}{\milli\meter}, the gantry length was adjusted slightly
to provide clearance between the grousers and the surrounding 2020 aluminum
profile. These changes did not alter the wheel radius, grouser count, or
sensing and control architecture used during testing.

To verify consistency between the two configurations, the
\SI{27.5}{\milli\meter} configuration was evaluated at an exposed height
of \SI{17.5}{\milli\meter}. The measured means remained within the bounds expected with the \SI{17.5}{\milli\meter} configuration at the same height, indicating comparable performance at the shared condition. {These trials were not included in the total of 950 fixed-height trials.}

For pea gravel and coarse rock, \SI{20.0}{\milli\meter} was included to
determine whether the deterioration above the \SI{7.0}{\milli\meter}
minimum continued beyond \SI{17.5}{\milli\meter}. A
\SI{22.0}{\milli\meter} condition was unnecessary because the minimum in
both terrains was already bounded. Sand required a broader range because no
reversal had occurred by \SI{17.5}{\milli\meter}; both packing states were
therefore evaluated at \SI{20.0}{\milli\meter},
\SI{22.0}{\milli\meter}, and \SI{24.0}{\milli\meter}.

\begin{figure}[!t]
    \centering
    \includegraphics[width=1\linewidth]{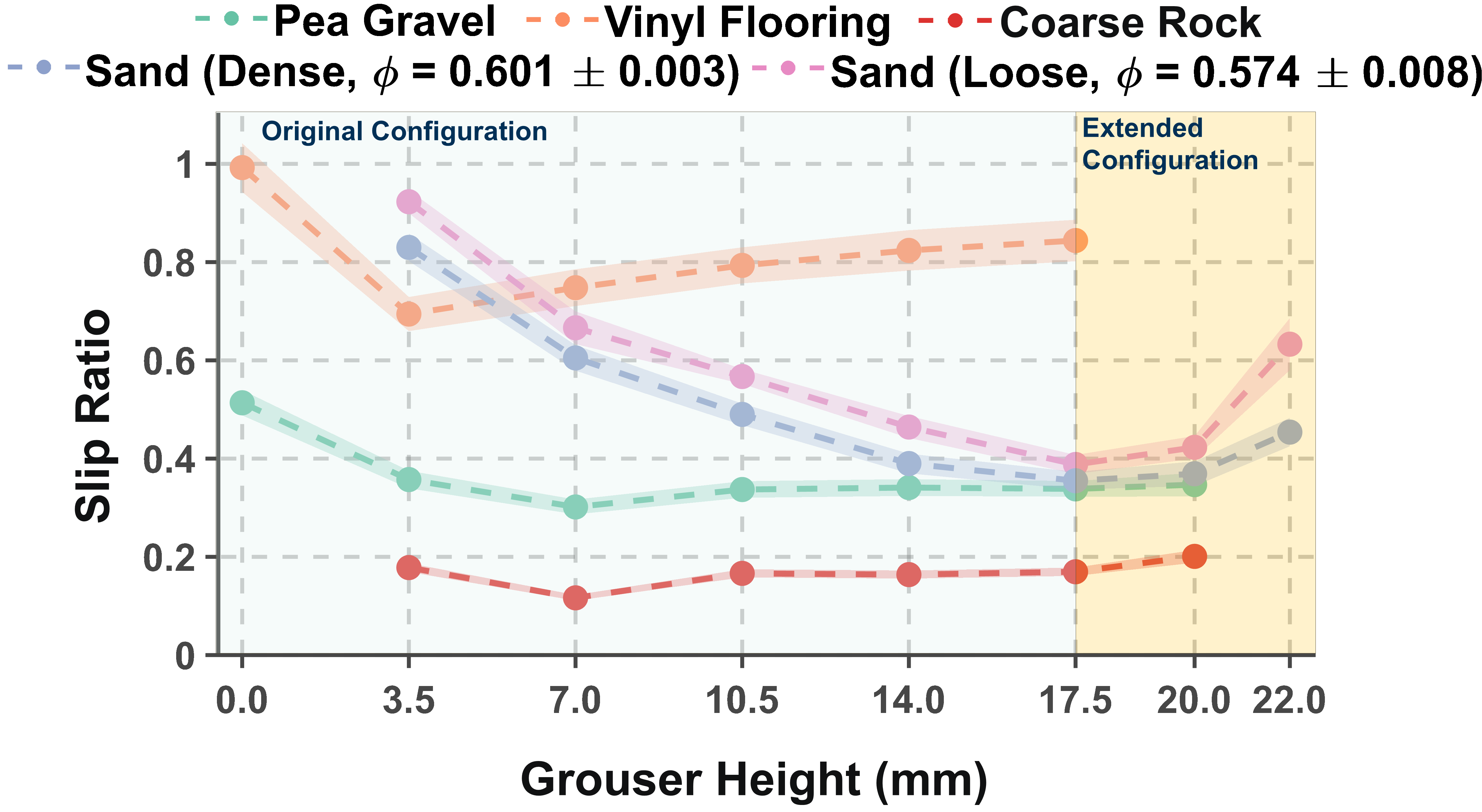}
    \caption{Slip ratio as a function of grouser height for four terrain
    types, including two sand volume-fraction states. The larger-height
    conditions bound the sand response while confirming continued
    degradation in pea gravel and coarse rock. Shaded regions represent
    $\pm 1$ standard deviation across trials.}
    \label{fig:r1}
\end{figure}

Twenty-five trials were performed for each larger-height terrain
configuration. The \SI{20.0}{\milli\meter} condition included 100 trials
across pea gravel, coarse rock, loose sand, and dense sand. The
\SI{22.0}{\milli\meter} and \SI{24.0}{\milli\meter} conditions each
included 50 trials across the two sand packing states. The complete
fixed-height characterization therefore consists of 950 trials across 38
terrain-height configurations. The resulting slip measurements are shown in \Cref{fig:r1}.

Across the common height range, the best tested configurations reduce slip
by approximately {34.6\%--58.0\%} relative to baseline values in the granular
terrains. On the non-granular vinyl, increasing height from
\SI{0.0}{\milli\meter} to \SI{3.5}{\milli\meter} reduces slip by 30.0\%
and decreases travel time by 97.6\%, but further increases in height hinder
performance. In pea gravel, \SI{7.0}{\milli\meter} grousers produce the
lowest measured slip of 0.302, reducing slip by 41.3\% relative to the
\SI{0.0}{\milli\meter} configuration. On coarse rock,
\SI{7.0}{\milli\meter} grousers similarly achieve the lowest measured slip
of 0.117, improving over the \SI{3.5}{\milli\meter} case of 0.178 by
34.6\%.

At larger heights, mean slip in pea gravel increases from 0.338 at
\SI{17.5}{\milli\meter} to $0.347 \pm 0.024$ at
\SI{20.0}{\milli\meter}. On coarse rock, mean slip increases from 0.170
to $0.201 \pm 0.013$ over the same height change. These results confirm
continued deterioration beyond \SI{17.5}{\milli\meter} for both terrains.

Sand shows a different response. In loose sand, raising the grousers from
\SI{3.5}{\milli\meter} to \SI{17.5}{\milli\meter} reduces slip by 58.0\%.
Mean slip then increases from 0.388 at \SI{17.5}{\milli\meter} to
$0.423 \pm 0.023$ at \SI{20.0}{\milli\meter} and
$0.633 \pm 0.053$ at \SI{22.0}{\milli\meter}. Densely packed sand follows
the same trend, increasing from 0.355 at \SI{17.5}{\milli\meter} to
$0.369 \pm 0.025$ at \SI{20.0}{\milli\meter} and
$0.454 \pm 0.029$ at \SI{22.0}{\milli\meter}. Among the discrete
configurations that complete the traversal, \SI{17.5}{\milli\meter}
therefore produces the lowest measured mean slip in both sand states. This
configuration is treated as the best tested height rather than an exact
continuous optimum.

Sand packing state also has a measurable effect on performance. Across the
common height range, dense packing reduces slip by approximately
{8.5\%--16.0\%} relative to the corresponding loose-sand cases. At larger
heights, this difference becomes more pronounced. At
\SI{20.0}{\milli\meter}, mean slip is 0.369 in dense sand compared with
0.423 in loose sand. At \SI{22.0}{\milli\meter}, the corresponding values
are 0.454 and 0.633, indicating greater degradation in the less densely
packed condition.

For both sand and coarse rock, GROVER is unable to complete
the testbed path at \SI{0.0}{\milli\meter}, as it either loses traction or
fails to climb over the terrain. In all such cases, 25 trials were still
conducted until immobilization. Together, the results show that no single
grouser height performs best across all terrains.

\begin{figure}[!t]
    \centering
    \includegraphics[width=1\columnwidth]{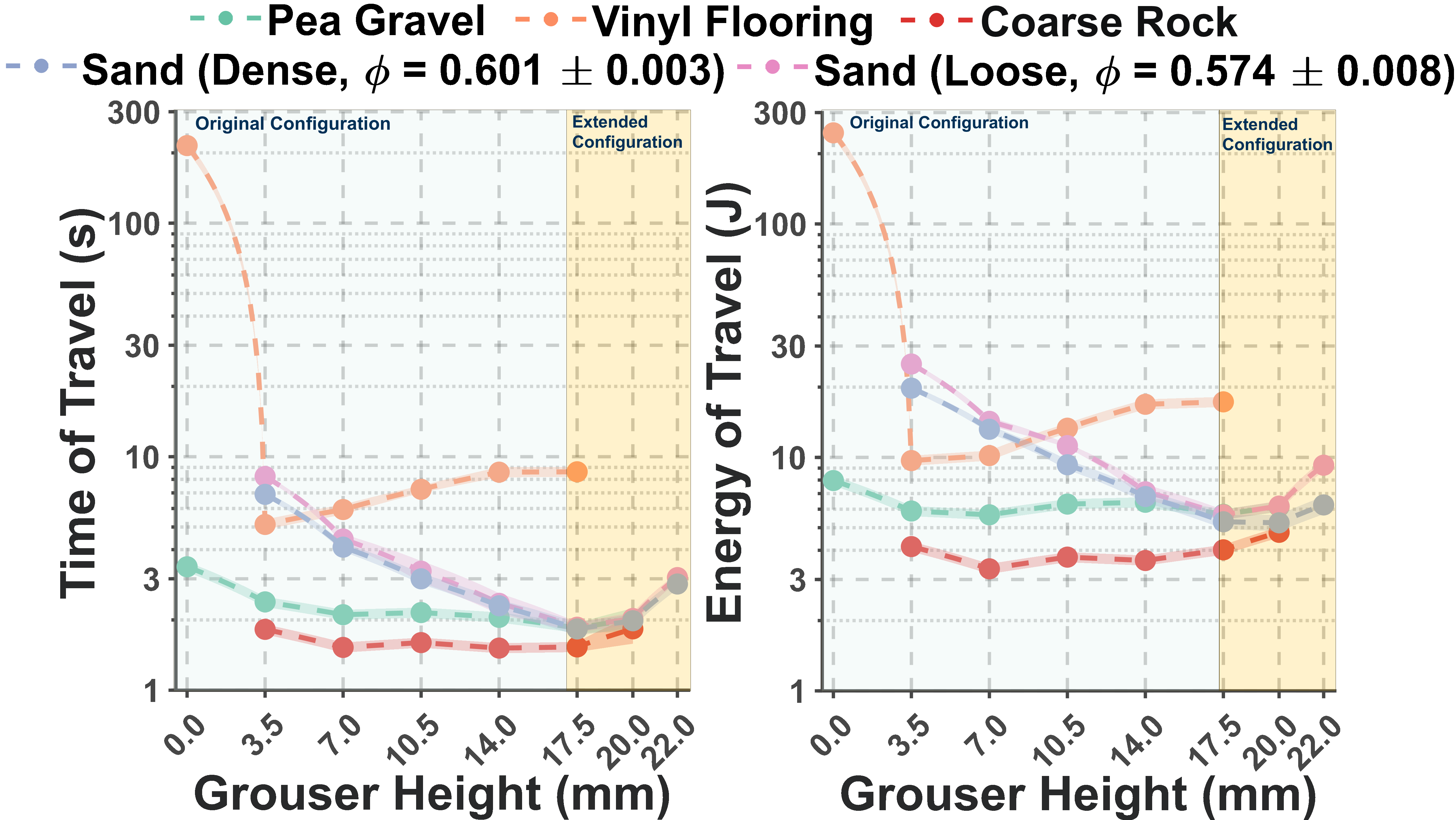}
    \caption{Travel time and energy of travel as functions of grouser
    height across four terrain types, including loose and densely packed
    sand. {The subplots span \SIrange{1}{300}{\second} and \SIrange{1}{300}{\joule}, respectively} and are
    displayed on a logarithmic scale to accommodate the wide dynamic
    range introduced by the \SI{0.0}{\milli\meter} vinyl configuration.
    Shaded regions represent $\pm 1$ standard deviation across trials.}
    \label{fig:r2}
\end{figure}

The energy and travel-time results are shown in \Cref{fig:r2}.

In loose sand, increasing height from \SI{3.5}{\milli\meter} to
\SI{17.5}{\milli\meter} reduces energy usage and travel time by 77.4\%.
In densely packed sand, the same height change reduces energy usage by
73.2\% and travel time by 73.5\%. In both sand states and on vinyl flooring,
energy consumption generally scales with travel time. While energy
consumption in pea gravel and coarse rock also follows travel time, the
relationship is less proportional, likely due to intermittent torque spikes
from larger granules.

In pea gravel, \SI{7.0}{\milli\meter} grousers reduce energy consumption
by 28.7\% and travel time by 37.6\% relative to the
\SI{0.0}{\milli\meter} case. In coarse rock,
\SI{7.0}{\milli\meter} grousers reduce energy consumption by 19.7\% and
travel time by 16.3\% relative to the \SI{3.5}{\milli\meter}
configuration.

{At the larger sand heights, loose-sand energy consumption and travel time
increase together with slip. Mean energy consumption and travel time increase
from \SI{5.677}{\joule} and \SI{1.862}{\second} at
\SI{17.5}{\milli\meter} to \SI{6.188}{\joule} and
\SI{2.029}{\second} at \SI{20.0}{\milli\meter}, and to
\SI{9.260}{\joule} and \SI{3.037}{\second} at
\SI{22.0}{\milli\meter}. In densely packed sand, travel time increases from
\SI{1.830}{\second} at \SI{17.5}{\milli\meter} to
\SI{1.984}{\second} at \SI{20.0}{\milli\meter} and
\SI{2.851}{\second} at \SI{22.0}{\milli\meter}. Mean energy changes only
from \SI{5.300}{\joule} to \SI{5.247}{\joule} between
\SI{17.5}{\milli\meter} and \SI{20.0}{\milli\meter}; this
\SI{0.053}{\joule} difference is small relative to the comparatively broad
$\pm 1$ standard-deviation band shown in \Cref{fig:r2}. Energy then increases
to \SI{6.257}{\joule} at \SI{22.0}{\milli\meter}. Thus, the
\SI{20.0}{\milli\meter} dense-sand condition exhibits longer travel time
without an increase in mean drive energy, whereas both metrics deteriorate at
\SI{22.0}{\milli\meter}.}

At \SI{24.0}{\milli\meter}, the wheel was unable to complete the traversal
in either loose or densely packed sand. These trials are therefore not
included as completed-traversal points in \Cref{fig:r1,fig:r2}. The
resulting immobilization behavior is examined separately in
\Cref{sec:sand_limit}.

\subsection{Sand Mobility Limit at Increased Grouser Height}
\label{sec:sand_limit}

\begin{figure}[!t]
    \centering
    \includegraphics[width=1\columnwidth]{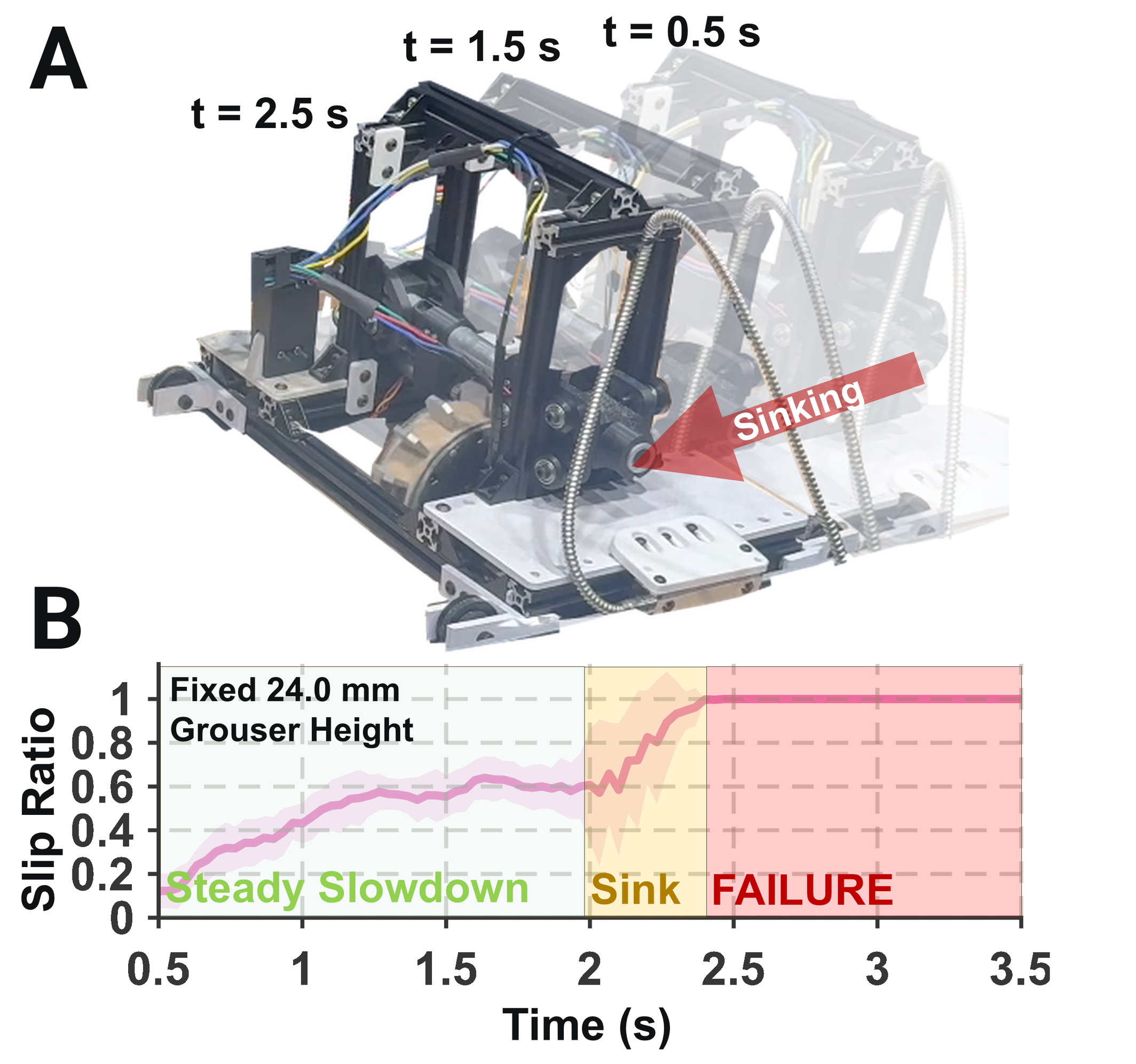}
    \caption{Wheel slowdown and immobilization with fixed
    {\SI{24.0}{\milli\meter}} grousers in loose sand
    ($\phi = 0.574 \pm 0.008$). (A) Overlaid video frames showing
    progression toward stall. (B) Mean slip ratio with $\pm 1$ SD
    shading. The initial transient is excluded, and the averaged window
    is truncated at the common analysis endpoint to avoid poorly
    supported late-time averages. Colored regions qualitatively
    illustrate steady motion, slowdown, sinkage, and failure and do not
    represent quantitatively defined states.}
    \label{fig:sinking}
\end{figure}

At the \SI{24.0}{\milli\meter} grouser-height condition, GROVER was unable to complete the traversal in either loose or densely packed sand across the 50 trials conducted at this height, with 25 trials performed for each packing state. \Cref{fig:sinking} presents the loose-sand condition as a representative case study, which was also documented in the recorded experimental media.

The time-resolved response shows a progressive loss of mobility prior to immobilization. Mean slip increases from approximately 0.12 at $t=\SI{0.5}{\second}$ to 0.43 at $t=\SI{1.0}{\second}$ and 0.55 at $t=\SI{1.5}{\second}$. The response remains near 0.60 through approximately $t=\SI{2.0}{\second}$ before increasing rapidly, reaching approximately 0.83 at $t=\SI{2.2}{\second}$ and 0.93 at $t=\SI{2.3}{\second}$. By approximately $t=\SI{2.4}{\second}$, mean slip approaches unity, corresponding to continued wheel rotation with effectively zero net translation. The early-time windowed slip is lower than the traversal-averaged slip measured at \SI{17.5}{\milli\meter}; however, these quantities are not directly equivalent because the former represents a local time window rather than an average over the complete traversal. The subsequent increase nevertheless shows that this initially lower slip is not sustained.

The variation between trials also increases substantially during the transition. Prior to $t=\SI{1.8}{\second}$, the slip standard deviation averages approximately 0.12, increasing to a maximum of approximately 0.34 near $t=\SI{2.03}{\second}$. This increased spread is primarily associated with trial-to-trial variation in the timing of slowdown and immobilization rather than increased steady-state variability. Small high-frequency translational fluctuations after immobilization are attenuated by the slip-processing window and smoothing.

{The recorded media also qualitatively show} greater accumulation of sand
between the \SI{24.0}{\milli\meter} grousers than was observed for
shorter configurations. This behavior was not quantified and is therefore
not used to assign a specific failure mechanism, but may motivate future
study of sand buildup or excavation effects at larger grouser heights.
For the present study, the \SI{24.0}{\milli\meter} condition is used only
as a case study demonstrating that increased grouser extension can
progress to complete mobility loss under the tested sand conditions.

\subsection{Terrain-Transition Demonstration}
\label{sec:terrain_transition}

The fixed-height characterization showed that pea gravel and loose sand
favor different grouser heights, motivating a test of whether the adaptive
wheel can reduce slip by changing its geometry during continuous traversal.
The wheel was therefore driven from pea gravel into loose sand using a
\SI{12}{\volt} wheel-motor supply at a fixed 85\% PWM duty cycle. The
reduced duty cycle provided additional time for grouser adjustment within
the available traversal distance. An initial grouser height of
\SI{12.0}{\milli\meter} was selected to demonstrate a short adjustment
during motion, rather than starting from the \SI{7.0}{\milli\meter}
configuration that produced the lowest slip in the fixed-height gravel
trials. This baseline reduced the commanded extension to the
\SI{17.5}{\milli\meter} sand setting from \SI{10.5}{\milli\meter} to
\SI{5.5}{\milli\meter}, a 47.6\% reduction in required displacement.

The height change was triggered at a traveled distance of approximately
\SI{247.8}{\milli\meter}. Because the gravel-to-sand boundary was
positioned manually, first sand contact varied slightly between trials.
Across 20 trials, transient slip peaks occurred at
{$t=1.95\pm0.10$~\si{\second}}, spanning
{\SIrange{1.80}{2.15}{\second}}, while the height command was
issued at {$t=2.10\pm0.06$~\si{\second}}. In 18 of 20 trials, the slip
peak occurred before the command, consistent with the visible slowdown
as the wheel entered the sand.

After the command, the measured grouser height reached 90\% of the
requested change in {$0.50\pm0.03$~\si{\second}}. Transient slip peaks
averaged $0.782\pm0.057$, while slip over
{$t=\SIrange{2.7}{3.3}{\second}$} averaged $0.436\pm0.016$, a
$44.0\pm4.7\%$ reduction from the transient peak. The later sand slip
remained approximately 32\% above the pre-transition gravel value.

\begin{figure}[!t]
    \centering
    \includegraphics[width=1\columnwidth]{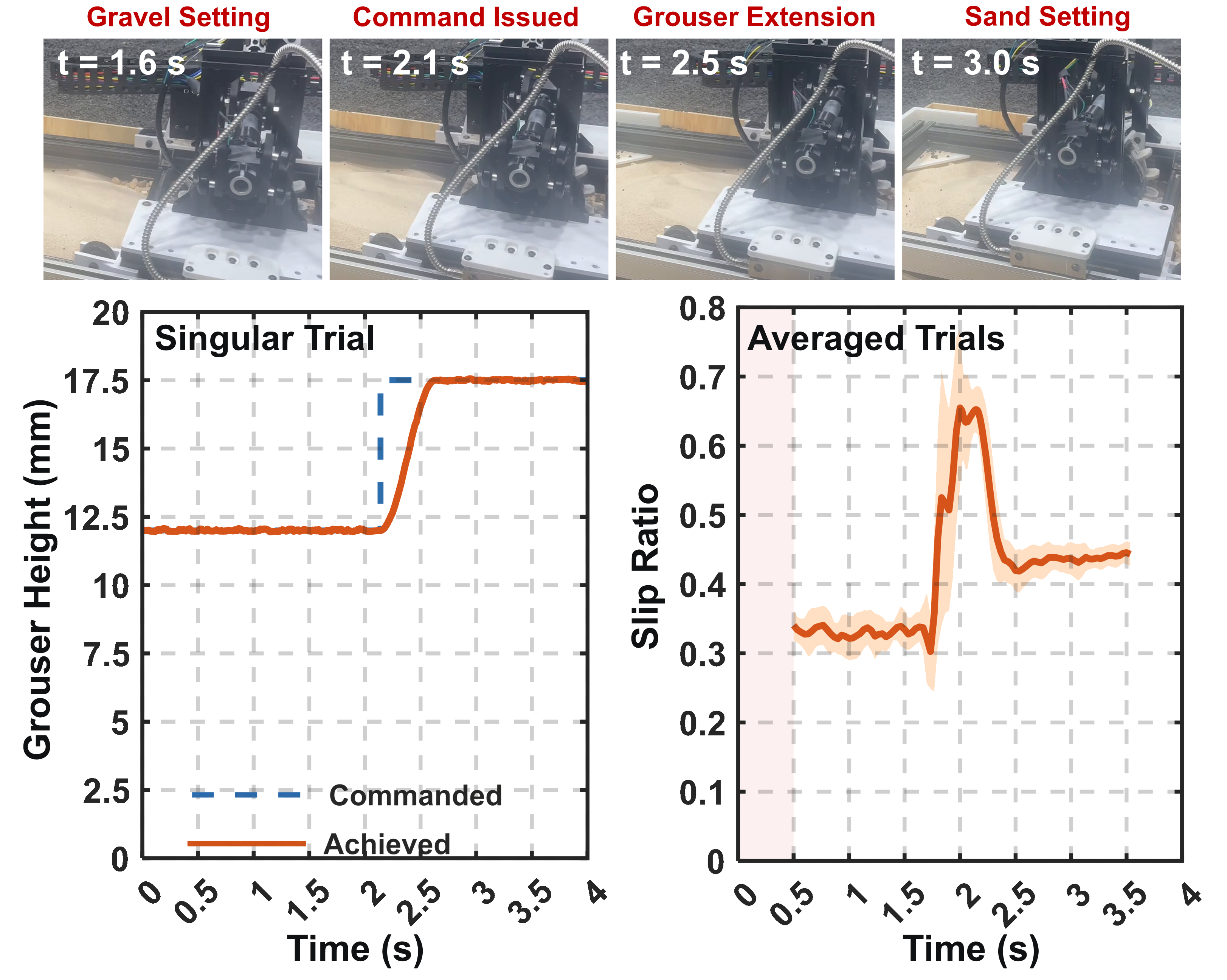}
    \caption{Distance-triggered grouser-height adjustment from
    \SI{12.0}{\milli\meter} to \SI{17.5}{\milli\meter} during transition
    from pea gravel to loose sand, using a \SI{12}{\volt} wheel-motor
    supply at 85\% PWM duty cycle. Video frames show the gravel setting,
    command issuance, grouser extension, and final sand setting. Left:
    commanded and achieved height for one representative trial. Right:
    20-trial mean slip ratio with $\pm 1$ SD shading. Trials are aligned
    by start time and truncated to the earliest common endpoint before
    averaging.}
    \label{fig:gravsand}
\end{figure}

The later sand slip was also higher than the mean value of 0.388 measured
at \SI{17.5}{\milli\meter} in the fixed-height loose-sand trials.
Although the loose-sand bulk packing state was maintained, local surface
homogeneity was less controlled in the manually assembled transition
terrain. Terrain entry and grouser extension also occurred within a short
traversal, so the later response may retain transient effects of changing
wheel--terrain contact and soil deformation. These effects could contribute
to the higher sand slip, but their individual contributions were not
quantified. Differences in drive duty cycle and averaging interval further
limit direct comparison with the fixed-height results.

Trials are aligned by start time rather than exact sand contact, so the
\SI{0.35}{\second} range in slip-peak timing contributes to the width of
the standard-deviation band during the transition. Slip is also calculated
over a finite sampling window and smoothed before averaging. The resulting
pointwise variation differs from the between-trial variation of
traversal-averaged slip in the fixed-height characterization. Truncation
to the earliest common endpoint maintains support from all 20 trials
throughout the plotted average.

This experiment shows that grouser height can be changed during continuous
traversal, with reconfiguration followed by a $44\%$ reduction from the
transient slip peak. On a rover, a camera or other terrain sensor could trigger the same height
change when a new surface is detected, allowing the wheel geometry to
adjust without stopping. A matched fixed-height transition was not performed, so the observed recovery does not isolate the effect of grouser adjustment from the terrain-entry transient.

\subsection{Online Adaptation Experiments}
\label{sec:online_experiments}

The terrain-transition experiment demonstrates grouser reconfiguration during
motion when the desired height is known in advance. The online adaptation
experiments test the next step: whether the wheel can select a useful grouser
height from slip feedback alone, without a terrain label or predefined
height lookup. The routine described in \Cref{fig:online} was evaluated on
four less-controlled surfaces: heterogeneous rock shards, grass-covered dirt,
compact soil, and AstroX's basalt regolith. Twenty-five trials
were performed on each terrain using a {\SI{12}{\volt}} motor supply at a $25\%$ duty
cycle. The reduced traversal speed provided sufficient time for the forward
search, reverse search, stability check, and final height command within the
available test length. These experiments evaluate whether measured slip can
drive useful grouser-height adaptation {across terrains with previously uncharacterized granularities};
they are not intended to demonstrate guaranteed convergence to a globally
optimal height.

Before formal testing, the search and convergence parameters in \Cref{tab:online_parameters} were selected using short preliminary traversals. During preliminary grass testing, the original stability tolerance was too restrictive for the larger slip fluctuations, prompting an increase in $\epsilon_0$ from \SI{0.050}{\per\second} to \SI{0.070}{\per\second}. This adjustment was applied globally, and the final parameters were held fixed across all four terrains throughout the 100 reported trials.

\begin{figure}[!t]
    \centering
    \includegraphics[width=1\columnwidth]{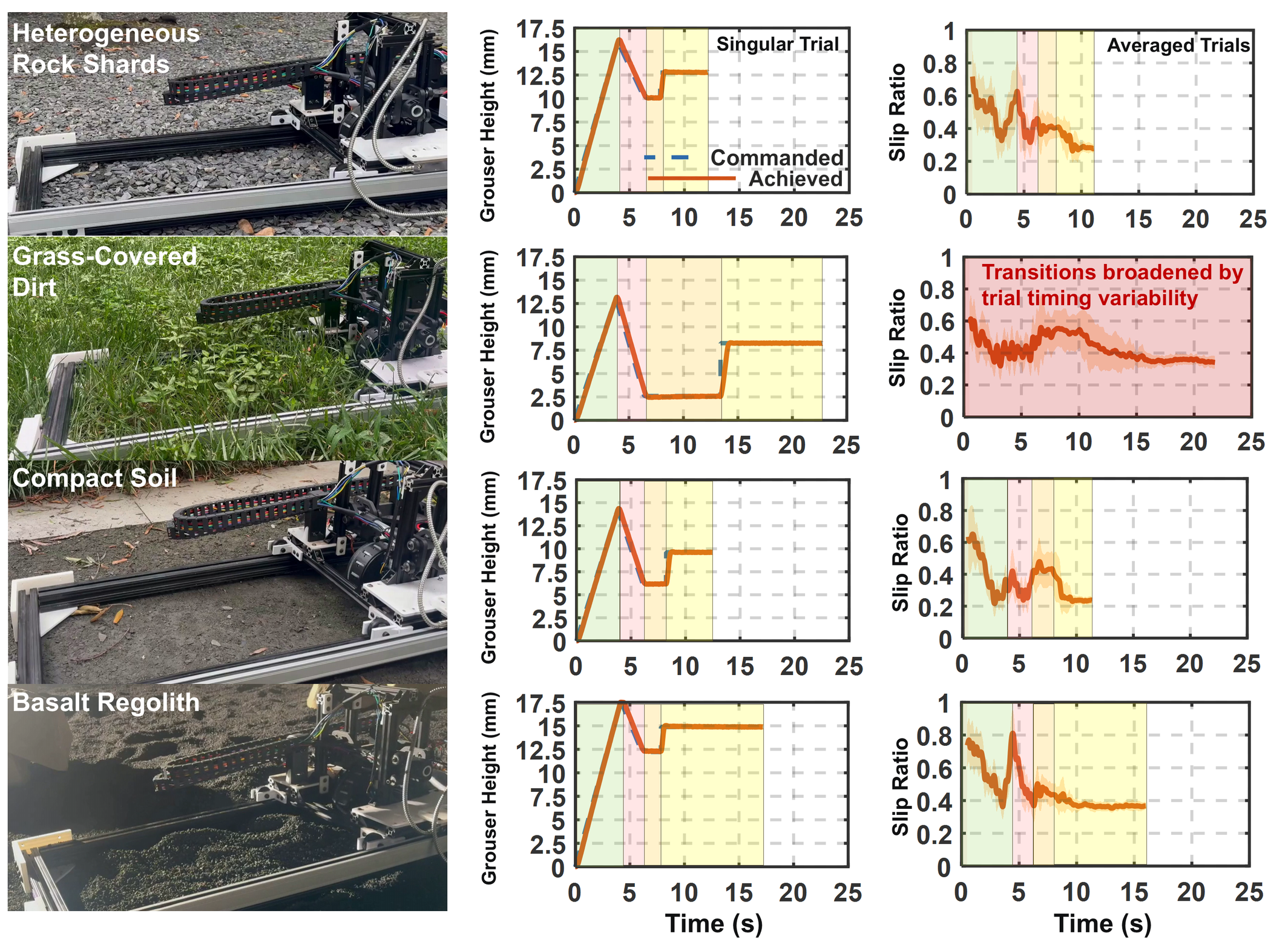}
    \caption{Online grouser-height adaptation on heterogeneous rock shards,
    grass-covered dirt, compact soil, and basalt regolith. The center panels
    show commanded and achieved grouser height for one representative trial
    on each terrain. Individual height trajectories vary in timing and shape,
    so averaging them would obscure the controller transitions and produce a
    trajectory not realized in any individual trial. The right panels show
    mean processed slip across 25 trials with shading corresponding to
    $\pm1$ sample standard deviation. Colored regions indicate mean controller
    phase timings, with broadened transitions showing trial-to-trial timing
    variation. The first {\SI{0.5}{\second}} is omitted to remove the initial wheel and
    filtering transient, and the traces are truncated to the common time
    interval for which all 25 trials contain valid data.}
    \label{fig:outside}
\end{figure}

All 25 trials on each terrain reached a final selected height. The selected
heights were {$13.26\pm0.66$~\si{\milli\meter}} on heterogeneous rock shards,
{$8.38\pm0.42$~\si{\milli\meter}} on grass-covered dirt, {$9.25\pm0.42$~\si{\milli\meter}} on compact soil,
and {$14.77\pm0.49$~\si{\milli\meter}} on basalt regolith. The different selected heights show
that the controller did not simply converge to a common intermediate
configuration across the four surfaces.

Further, the timing of the final height command was less consistent. It occurred at
{$8.29\pm0.37$~\si{\second}} on rock shards, {$11.99\pm1.78$~\si{\second}} on grass,
{$8.13\pm0.33$~\si{\second}} on compact soil, and {$8.36\pm0.43$~\si{\second}} on basalt. Grass was
the clear outlier, with final-command times ranging from {\SIrange{9.74}{15.24}{\second}}. By comparison, the time from the final command to mechanical
settling remained relatively consistent at {$0.74\pm0.08$~\si{\second}}, {$0.88\pm0.11$~\si{\second}},
{$0.73\pm0.08$~\si{\second}}, and {$0.78\pm0.12$~\si{\second}}, respectively. The larger timing spread
on grass therefore results primarily from the slip-based search and stability
logic rather than the grouser-height response itself.

To characterize the overall change in slip without selecting a single
transient maximum, the mean processed slip from {$t=\SI{0.5}{\second}$} until final height
settlement was compared with the mean slip after settlement for each trial.
The pre-settled and settled values were $0.450\pm0.083$ and
$0.284\pm0.028$ on rock shards, $0.480\pm0.075$ and $0.351\pm0.026$ on
grass, $0.387\pm0.071$ and $0.236\pm0.020$ on compact soil, and
$0.527\pm0.067$ and $0.368\pm0.020$ on basalt. Calculated within each trial,
this corresponds to mean reductions of $35.6\pm9.0\%$, $25.4\pm10.9\%$,
$37.2\pm11.5\%$, and $29.1\pm9.8\%$, respectively. The settled mean was
lower than the pre-settled mean in all 25 trials on each terrain.

These reductions describe the full adaptation sequence rather than the
isolated effect of grouser height. The wheel continues to move across spatially
varying terrain while the morphology changes, so local terrain variation also
affects the measured slip. The controller also does not consistently retain
the lowest transient slip observed during a search, and no exhaustive
fixed-height sweep was performed on these outdoor surfaces to identify a
ground-truth optimum.

The transient responses further show this limitation. On compact soil, slip
increases after the reverse-search command stops before the final configuration
settles, whereas rock shards and basalt remain comparatively stable over the
same phase. Grass shows substantially larger fluctuations and phase-timing
variation, making parts of its time-averaged response difficult to interpret.
Because the trials are averaged by elapsed time rather than aligned at each
controller transition, the standard-deviation bands reflect both slip
variability and differences in when individual trials enter each phase.

The primary contribution of these experiments is the demonstration that
grouser-height adaptation can reduce measured slip {during traversal across terrains with previously uncharacterized granularities}. Across all four surfaces, the
controller used slip feedback alone to search the available height range,
select a final configuration, and reach a lower-slip settled operating state
than during the preceding search. The selected heights also differed between
terrains, indicating that the controller responded to the measured mobility
conditions rather than converging to a fixed configuration. However, the
method does not consistently identify the minimum-slip configuration, and the
timing variability and the fact that grass needed a new convergence parameter after initial tuning remain important
limitations of the present approach.

\section{Conclusion}
\label{sec:conclusion}

This work introduced GROVER, a continuously variable grouser-height wheel designed to improve mobility by adapting wheel geometry to terrain conditions. Fixed-height characterization across 950 trials and 38 terrain-height configurations showed that the best tested grouser height varies with terrain. Across the granular terrains, the best configurations reduced slip by {34.6\%--58.0\%} and improved travel time and energy consumption by up to 77.4\%. No single grouser height minimized slip across all terrains, demonstrating the limitation of selecting one fixed configuration for variable terrain.

The platform was then evaluated while changing grouser height during traversal. During an instructed terrain transition, reconfiguration was followed by a $44.0\pm4.7\%$ reduction from the transient slip peak across 20 trials. The online adaptation routine extended this capability {across terrains with previously uncharacterized granularities} by using slip feedback alone to search the available grouser-height range. Across three field terrains and one laboratory regolith terrain, the controller selected different final heights, with settled slip {25.4\%--37.2\%} lower than the pre-settled mean across 25 trials per terrain.

These results demonstrate the central capability of GROVER: grouser height can be treated as an active mobility parameter and varied during traversal to reach lower-slip operating states. The instructed and online experiments represent complementary approaches, with one using a known terrain-dependent configuration and the other selecting a configuration directly from measured mobility feedback. These findings suggest that terrain-responsive wheel morphologies can expand the operational envelope of rovers in mobility-challenging extraterrestrial environments.

\bibliographystyle{IEEEtran}
\bibliography{references} 

\section*{Conflict of Interest}
The authors declare that U.S. Provisional Patent Application No. 63/903,602, related to this work, was filed on 22 October 2025.

\end{document}